\pdfoutput=1
\documentclass[11pt]{article}

\usepackage[preprint]{acl}
\usepackage{float}
\usepackage{subcaption}
\usepackage{amsmath}
\usepackage{amssymb}
\usepackage{times}
\usepackage{latexsym}
\usepackage[normalem]{ulem}
\usepackage{algorithm}
\usepackage{algorithmic}
\usepackage[T1]{fontenc}
\usepackage[utf8]{inputenc}

\usepackage{microtype}

\usepackage{inconsolata}

\usepackage{graphicx}
\usepackage{float}

\usepackage{multirow} 
\usepackage{booktabs}       % professional-quality tables
\usepackage{amsmath}        % for \text in math mode
\usepackage{amsfonts}       % blackboard math symbols
\usepackage{xcolor}         % colors (needed for the change marks below)
\title{
Sparse Subspace-to-Expert Sharing for Task-Agnostic Continual Learning}
\author{
\textbf{Fatema Siddika}\textsuperscript{1},
\textbf{Md Anwar Hossen}\textsuperscript{1},
\textbf{Tanwi Mallick}\textsuperscript{2},
\textbf{Ali Jannesari}\textsuperscript{1} \\
\textsuperscript{1}Iowa State University, Ames, USA \\
\textsuperscript{2}Argonne National Laboratory, USA \\
\texttt{\{fatemask, manwar, jannessar\}@iastate.edu} \\
\texttt{tmallick@anl.gov}
}
\begin{document}
\maketitle
\begin{abstract}
Continual learning in Large Language Models (LLMs) is hindered by the plasticity-stability dilemma, where acquiring new capabilities often leads to catastrophic forgetting of previous knowledge. Existing methods typically treat parameters uniformly, failing to distinguish between specific task knowledge and shared capabilities. We introduce Mixture of \textbf{S}parse \textbf{E}xperts for \textbf{T}ask \textbf{A}gnostic Continual Learning (SETA), a framework that resolves the plasticity-stability conflict through adaptive sparse subspace decomposition into task-specific expert modules. Unlike standard updates, where tasks compete for the same parameters, SETA separates knowledge into unique experts, designed to isolate task-specific patterns, and shared experts, responsible for capturing common features. SETA shares an overlapping parameter block only when the new task's update passes a subspace-agreement test and a sign test against its prior owner, shared knowledge is protected by adaptive elastic anchoring, and independently gated routers retrieve the relevant experts at inference without task identifiers. Extensive experiments across diverse domain-specific benchmarks demonstrate that SETA achieves competitive or superior overall performance relative to state-of-the-art continual learning baselines, improving overall performance over the strongest baseline by 1.1 to 1.7 points on TRACE and by up to 5.5 points on the O-LoRA benchmark, with backward transfer between $-2.2$ and $-1.3$ on TRACE, against 4.6 to 6.3 points of forgetting for MoE-CL and 8 to 15 points for the other LoRA baselines. The paper code is available at \href{https://github.com/FatemaSiddika/Split-on-Share}{https://github.com/FatemaSiddika/Split-on-Share}.
\end{abstract}
\section{Introduction}
Large Language Models have demonstrated remarkable performance across diverse natural language processing tasks, ranging from text classification and reasoning to dialogue and summarization~\cite{brown2020language, raffel2020exploring}. However, deploying LLMs in continual learning settings remains challenging~\cite{parisi2019continual} as models must adapt to sequential tasks while preserving previously acquired knowledge~\cite{kirkpatrick2017overcoming, li2017learning}. This challenge, known as catastrophic forgetting~\cite{li2017learning, kirkpatrick2017overcoming}, arises from the fundamental trade-off between learning plasticity, the rapid adaptation to new tasks, and memory stability, the retention of learned knowledge. The problem is further exacerbated in parameter-efficient fine-tuning regimes, where sparsity for efficiency increases susceptibility to overwriting historical information~\cite{han2015learning, evci2020rigging}.

Existing continual learning strategies generally fall into three primary categories, including Replay based methods that rehearse historical data~\cite{chaudhry2019tiny}, Regularization based techniques that penalize updates to critical parameters~\cite{kirkpatrick2017overcoming}, and Parameter Isolation approaches that physically separate task specific modules~\cite{rusu2016progressive}. Despite their differences, these approaches share a common limitation: they treat parameters uniformly within their respective subspaces~\cite{ren2024analyzing}.
To address these limitations, Continual Sparse Fine-Tuning provides a natural mechanism for uncovering the internal knowledge structure of large language models. Unlike parameter-efficient fine-tuning (PEFT)-based low-rank adapters that project updates through opaque dense matrices, sparse tuning identifies specific attention sub-blocks, particularly in the Value projection, based on high-utility gradient signals. This unique property serves as the foundation for our proposed framework, Mixture of Sparse Experts for Task Agnostic Continual Learning, referred to as \textbf{SETA}.
The design of SETA is driven by three fundamental research questions.
\par \textit{\textbf{RQ1:} How can sparse parameter patterns be leveraged to decouple the acquisition of novel features from the retention of historical representations?}
To address RQ1, SETA introduces a subspace-based expert decomposition that frames sparse continual learning within a Mixture-of-Experts paradigm. By exploiting the inherent block-wise sparsity of gradients, the model is decomposed into distinct expert modules, ensuring that conflicting updates operate in isolated subspaces rather than competing for overlapping parameter blocks.
\par \textit{\textbf{RQ2:} Is it possible to automatically differentiate between reusable shared knowledge and task-specific unique knowledge within a single unified framework?}
To answer RQ2, the Split-on-Share (SoS) mechanism shares an overlapping parameter block only when the new task's update passes a subspace-agreement test and a sign test against its prior owner; such blocks form shared experts, protected by adaptive elastic anchoring that scales with each expert's task history and drift. All other parameters become unique experts, which are frozen once their task ends to preserve historical knowledge.
\par \textit{\textbf{RQ3:} How can the relevant expert modules be dynamically retrieved during inference without relying on external task identifiers?} SETA resolves RQ3 by implementing adaptive gating expansion with logit invariance, which preserves existing routing decision boundaries during expert splitting, together with independent per-expert sigmoid gating that keeps each expert's activation decoupled from the rest of the population.
\par We evaluate SETA on TRACE with three LLM backbones and on the standard and long-sequence orders of the O-LoRA
benchmark with Qwen3-4B. On TRACE-8, SETA achieves the highest overall performance (OP) and the least forgetting
(BWT), improving OP by 1.1 to 1.7 points over the strongest baseline and keeping BWT between $-2.2$ and
$-1.3$.
\section{Related Work}
Parameter-efficient fine-tuning (PEFT) adapts large language models by updating a small set of parameters while the pre-trained weights stay frozen~\cite{li2021prefix}. A variety of PEFT techniques have been proposed, including Adapter Tuning~\cite{houlsby2019parameter}, BitFit~\cite{zaken2022bitfit}, Prefix Tuning~\cite{li2021prefix}, Prompt Tuning~\cite{lester2021power}, and Low-Rank Adaptation (LoRA)~\cite{hu2022lora}. Extensions such as ReLoRA~\cite{lialin2024relora} and RankAdapter~\cite{zhou2024rankadaptor} improve memory efficiency and dynamically adjust ranks, though they lack formal guarantees. AdaZeta~\cite{yang2024adazeta} introduces zeroth-order optimization with convergence guarantees, while other methods~\cite{gao2024adaptive} explore adaptive-rank strategies without theoretical proofs. LoRA has also been combined with Mixture-of-Experts architectures~\cite{li2024mixlora, huang2024mixture}, as in AdaMoLE~\cite{liu2024adamole}, enabling dynamic expert selection, and with Neural Architecture Search for LLM compression~\cite{munoz2025low}. Another line of work leverages sparsity, such as Sparse Matrix Tuning (SMT)~\cite{he2025smt}, which selects task-relevant sub-matrices.    \citet{anonymous2026fedseam} extends sparse expert adaptation to federated continual learning, aggregating clients' experts on a server. CRAFT~\citep{hossen2026craft} adapts hidden representations rather than weights.
\section{Problem Formulation and Motivation}
\subsection{Continual Learning Setting}
Continual learning aims to enable models to acquire new capabilities 
from a sequence of tasks $\mathcal{T} = \{T_1, \dots, T_k\}$ without 
forgetting previously learned knowledge~\cite{wu2024continual}. 
To address this, parameter-efficient fine-tuning (PEFT) methods such as 
LoRA freeze the backbone and train small low-rank modules~\cite{hu2022lora}. 
However, restricting updates to a narrow bottleneck dimension ($r \ll d$) 
creates a geometric disconnect between sequential task minima~\cite{ren2024analyzing}, 
introducing noise that hinders knowledge retention over long task 
sequences.
\begin{figure*}[htbp]
    \centering
    \includegraphics[width=1.02\linewidth]{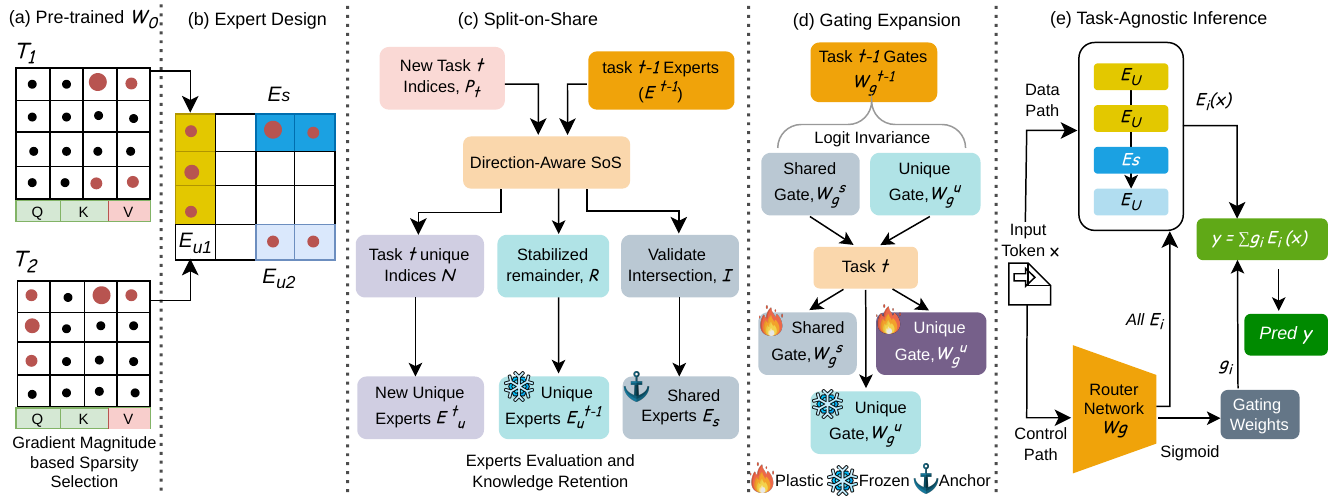}
    \caption{Overview of SETA. (a--b) High-utility V-projection blocks are selected from the pretrained LLM by
gradient scores. (c) Split-on-Share turns overlapping blocks that pass the direction test into plastic shared
experts ($E_{\text{s}}$) and keeps the remaining blocks in frozen unique experts ($E_{\text{u}}$). (d) The gating
network grows with logit invariance, so routing decisions are preserved when an expert splits. (e) At inference,
independent sigmoid gates weight all experts for each token, without task identifiers.}
\label{fig:seta_architecture}
    \label{fig:seta_diagram}
\end{figure*}
\subsection{Shared Subspace Dilemma}
The core failure of continual learning arises from the Shared Subspace Dilemma. Let $\mathcal{P}_t$ denote parameters for task $t$; its intersection with the cumulative historical subspace $\mathcal{P}_{1:t-1} = \bigcup_{k=1}^{t-1} \mathcal{P}_k$ defines a shared region $\mathcal{I} = \mathcal{P}_t \cap \mathcal{P}_{1:t-1}$. Here, updating weights risks overwriting prior knowledge, while freezing limits plasticity. Standard methods address this via uniform regularization~\cite{kirkpatrick2017overcoming}:
\begin{equation}
    \mathcal{L}_{\text{reg, uniform}} = \gamma \cdot \| W - W^{*}_{1:t-1} \|^2,
\end{equation}
where $\gamma$ is a fixed penalty. However, uniform regularization is too rigid, treating parameters equally rather than distinguishing shared features requiring controlled plasticity from task-specific regions needing stability~\cite{lopez2017gradient}. Resolving this trade-off requires both a structural solution to isolate parameter subspaces 
and an adaptive regularization strategy.
\section{Methodology}
\label{sec:method}
SETA introduces Split-on-Share (Section~\ref{sec:split_on_share}),
adaptive elastic anchoring (Section~\ref{sec:knowledge_preservation}) and logit-invariant gating with independent
sigmoid gates (Sections~\ref{sec:gating_expansion}--\ref{sec:sigmoid_gating}); Algorithm~\ref{alg:seta_algo}
(Appendix~\ref{app:algorithm}) gives the full procedure.

\subsection{Sparse Subspace Selection}
\label{sec:sparse_construction}
SETA fine-tunes only task-relevant sparse sub-matrices (Figure~\ref{fig:seta_architecture}a); a subspace here is a
block-sparse optimization boundary found from gradients, not a semantically disentangled latent space. We partition
each pre-trained weight matrix $W \in \mathbb{R}^{d \times m}$ into a grid of $l \times l$ blocks
$\mathbb{B} = \{B_{i,j}\}$, with $l$ dividing the projection exactly ($l = 256$ for LLaMA-2-7B, $l = 128$ for
LLaMA-3-8B and Qwen3-4B) to avoid remainder blocks and keep memory access contiguous. Blocks are scored by a parameter-free pipeline during a warm-up phase before each task.\\
\textbf{Dynamic Warm-up Stopping:} We monitor the per-step gradient contribution ratio
\begin{equation}
    \rho(t) = \frac{\bigl|\;\|\nabla W\|_t - \|\nabla W\|_{t-1}\;\bigr|}{\|\nabla W\|_t},
\end{equation}
where $\|\nabla W\|_t$ is the accumulated gradient norm after step $t$. Warm-up halts once $\rho(t) \leq \mu_\rho + \sigma_\rho$, estimated over
a trailing window of $W$ steps, holds for $S$ consecutive steps, with a larger patience $S$ for generation tasks
($W$, $S$ and the step bounds are in Appendix~\ref{app:datasets}).\\
\textbf{Layer-Normalized Scoring:}
The raw importance of a block is its mean absolute gradient:
\begin{equation}
    \mathcal{S}_{i,j}^{(k)} = \frac{1}{l^2} \sum_{(u,v) \in B_{i,j}} |\nabla W_{u,v}|.
\end{equation}
To keep dominant layers and outlier input channels from monopolizing the global ranking, each layer's score
matrix is standardized per input-column block and then to zero mean and unit variance, with negative scores set to
zero.\\
\textbf{Cross-Task Divergence Scoring:}
For tasks $t > 1$, the normalized score $\hat{\mathcal{S}}_t$ is augmented with a divergence signal rewarding
blocks with a novel gradient response, via the convex combination
\begin{equation}
    \tilde{\mathcal{S}}_t = \alpha \cdot \hat{\mathcal{S}}_t +
    (1-\alpha) \cdot \left|\hat{\mathcal{S}}_t - \bar{\mathcal{S}}_{1:t-1}\right|,
\end{equation}
where $\bar{\mathcal{S}}_{1:t-1}$ is the mean block score over all prior tasks and $\alpha = 1/(1 + \log t)$ decays
with the task count, introducing no additional hyperparameter.\\
\textbf{Adaptive Block Budget:}
The block budget is a fraction $\beta$ of all blocks, analogous to a LoRA rank, giving
$N_{\text{base}} = \lfloor \beta \cdot |\mathbb{B}| \rfloor$, with per-layer quotas inversely proportional to
historical coverage, $q(l) \propto 1/(c(l)+\varepsilon)$. Blocks are filled in score order within these quotas,
excluding blocks already selected by two earlier tasks, which steers later tasks toward unused capacity. Since
90-94\% of high-magnitude gradients concentrate in the Value ($V$) projection (Figure~\ref{fig:gradient_analysis},
Appendix~\ref{app:phase0:overlap}), attributed to softmax saturation where $Q$ and $K$ gradients vanish as
$QK^\top / \sqrt{d_k}$ grows, we restrict selection to the V-projection.

\subsection{Unique and Shared Experts}
\label{sec:experts}
Shared experts $E_{\text{s}}$ hold the part of the overlap $\mathcal{P}_t \cap \mathcal{P}_{1:t-1}$ between the
selected and historical subspaces that passes the direction test (Section~\ref{sec:direction_sharing},
Figure~\ref{fig:seta_architecture}b); the unique expert $E_{\text{u}}^{(t)}$
holds the rest of $\mathcal{P}_t$. The effective weight is
\begin{equation}
\label{eq:effective_weight}
    W = W_0 + \sum_{j \in \mathcal{E}_{\text{u}}} \Delta W_j + \sum_{i \in \mathcal{E}_{\text{s}}} \Delta W_i.
\end{equation}

\subsection{Split-on-Share (SoS) Expert Evolution}
\label{sec:split_on_share}
\label{sec:direction_sharing}
We propose Split-on-Share (SoS), which carves shared experts from overlapping blocks that pass a direction test, under
one layer-wise size rule (Figure~\ref{fig:seta_architecture}c).\\
\textbf{Direction-Aware Sharing:}
Positional overlap alone does not show that two tasks use a block alike, so we adopt the direction test of \citet{anonymous2026fedseam}. Let $D_{\text{old}}$ be the prior expert's trained delta on block $B_{i,j}$ and
$G_{\text{new}}$ the new task's gradient there, measured while the pretrained model is briefly fine-tuned on the
new task. With $Q_{\text{old}}$ and $Q_{\text{new}}$ their top right singular vectors (90\% of the energy, rank at
most 8), the score
\begin{equation}
    s_{i,j} = \frac{\bigl\| Q_{\text{old}}^\top Q_{\text{new}} \bigr\|_F^2}{\min(r_{\text{old}}, r_{\text{new}})}.
\end{equation}
is the mean squared principal cosine of the two input subspaces. The block passes if $s_{i,j}$ exceeds the mean
plus $Z$=3 standard deviations of the scores of $D_{\text{old}}$ against the new task's gradients on up to 32 blocks
it selected in the same layer (random subspaces of equal rank when there are too few). Since $s_{i,j}$ ignores
sign, the block must also satisfy
\begin{equation}
    c_{i,j} = \frac{\langle D_{\text{old}}, -G_{\text{new}} \rangle}
    {\|D_{\text{old}}\|_F \, \|G_{\text{new}}\|_F} > \tau,
\end{equation}
with $\tau=0$ by default, a first-order condition: a small step along $D_{\text{old}}$ lowers the new task's loss near the pretrained
weights, while the prior task's loss is not checked. A block that fails either test stays frozen in the prior
expert, and the new task trains its own zero-initialized copy, whose gradient is projected off the span of
$Q_{\text{old}}$; under Adam this steers the copy away from those directions but does not keep it exactly
orthogonal.\\
\textbf{Minimum Expert Size:}
For each layer $l$, let $\mathcal{I}_l$ be the validated intersection,
$\mathcal{R}_l = \mathcal{P}_{t-1}^{(l)} \setminus \mathcal{I}_l$ the remainder of the prior expert, and
$\mathcal{N}_l = \mathcal{P}_{t}^{(l)} \setminus \mathcal{I}_l$ the new task's own indices. Every expert piece must
reach a minimum size $T = \varphi \cdot N_{\text{base}}$, tied to the block budget rather than set as an absolute
count, so a layer is carved only if
\begin{equation}
\label{eq:sos_min_size}
    |\mathcal{I}_l| \geq T, \qquad |\mathcal{R}_l|,\, |\mathcal{N}_l| \in \{0\} \cup [T, \infty);
\end{equation}
otherwise $\mathcal{I}_l \leftarrow \emptyset$, the prior expert keeps all its blocks, and the new task keeps its
own copy of the overlap. Layers in which the new task selects fewer than $T$ blocks are not adapted.\\
\textbf{Expert Formation:}
$\mathcal{I}_l$ instantiates the plastic shared expert $E_{\text{s}}$ by weight inheritance,
$W_{\text{s}} \leftarrow W_{t-1}[\mathcal{I}_l]$; the residuals $\mathcal{R}_l$ form the frozen unique expert
$E_{\text{u}}^{(t-1)}$; and $\mathcal{N}_l$ initializes the new unique expert $E_{\text{u}}^{(t)}$ from the base weights.

\subsection{Knowledge Retention and Training Objective}
\label{sec:knowledge_preservation}
\textbf{Subspace Isolation:}
While task $t$ is trained, all prior unique experts are frozen by a zero-gradient constraint,
\begin{equation}
    \forall w \in \bigcup_{k < t} \mathcal{E}_{\text{u}}^{(k)}, \quad \frac{\partial \mathcal{L}}{\partial w} \equiv 0,
\end{equation}
so only the shared experts $E_{\text{s}}$, the current unique expert $E_{\text{u}}^{(t)}$ and the gating networks
are optimized; the gate rows of all experts, including earlier ones, stay trainable
(Section~\ref{sec:gating_expansion}).\\
\textbf{Adaptive Elastic Anchoring:}
To protect shared experts from semantic drift without blocking forward transfer, we propose a per-expert
coefficient that adapts at every training step $s$. Let $\hat{W}_i$ be the snapshot of shared expert $i$ taken
before task $t$ begins. The accumulation weight $\omega_i = n_{\text{prior}}^{(i)} / (t-1)$ scales protection with
the number of prior tasks in expert $i$, and the drift factor tracks the displacement from the snapshot:
\begin{equation}
    \delta_i(s) = 1 + \tanh\!\left(\frac{\|W_i^{(s)} - \hat{W}_i\|}
    {\max(\|\hat{W}_i\|,\,\epsilon)}\right),
\end{equation}
yielding the coefficient
\begin{equation}
    \lambda_i(s) = \lambda_{\text{base}} \cdot \omega_i \cdot \delta_i(s).
\end{equation}
Here $\delta_i(s) \in [1, 2)$ is non-decreasing as the expert drifts ($\delta_i=1$ when $W_i^{(s)}=\hat{W}_i$). The
regularization loss is
\begin{equation}
   \mathcal{L}_{\text{reg}} = \sum_{i \in \mathcal{E}_{\text{s}}}
    \lambda_i(s) \cdot \|W_i^{(s)} - \hat{W}_i\|^2.
\end{equation}
\textbf{Training Objective:} For $T_1$ we minimize the standard sparse fine-tuning loss; for $t > 1$,
\begin{equation}
\label{equ:seta_train_object}
        \mathcal{L} = -\sum_{(x,y)\sim \mathcal{D}_t} \log P(y \mid x; \Theta_t) + \mathcal{L}_{\text{reg}},
\end{equation}
where $\Theta_t$ contains the expert weights and the gate parameters at task $t$, combined as in Eq.~\eqref{eq:task_agnostic_pred}.
\begin{table*}[t]
\centering
\caption{Comparison of continual learning methods on TRACE-8 with three backbones. We report OP and BWT (\%) after
training on the full sequence as mean$_{\pm\text{std}}$ over two seeds. FIX prompts the frozen backbone with
in-context examples, so it has no BWT. SETA uses $\beta$=12\%. \textbf{Bold} marks the best mean in each column.}
\label{tab:all_models_summary}
\small
\setlength{\tabcolsep}{4pt}
\begin{tabular}{lcccccc}
\toprule
& \multicolumn{2}{c}{Qwen3-4B} & \multicolumn{2}{c}{LLaMA-2-7B} & \multicolumn{2}{c}{LLaMA-3-8B} \\
\cmidrule(lr){2-3}\cmidrule(lr){4-5}\cmidrule(lr){6-7}
Method & OP $\uparrow$ & BWT $\uparrow$ & OP $\uparrow$ & BWT $\uparrow$ & OP $\uparrow$ & BWT $\uparrow$ \\
\midrule
FIX & $46.55_{\pm 0.72}$ & -- & $30.29_{\pm 1.86}$ & -- & $36.50_{\pm 0.11}$ & -- \\
SeqLoRA & $45.71_{\pm 3.29}$ & $-7.97_{\pm 1.83}$ & $41.97_{\pm 0.84}$ & $-9.72_{\pm 0.13}$ & $46.92_{\pm 2.66}$ & $-12.32_{\pm 3.12}$ \\
O-LoRA & $45.13_{\pm 1.94}$ & $-9.93_{\pm 1.06}$ & $40.32_{\pm 1.93}$ & $-11.76_{\pm 2.11}$ & $49.54_{\pm 0.96}$ & $-9.14_{\pm 2.55}$ \\
HiDe-LoRA & $43.73_{\pm 0.77}$ & $-9.35_{\pm 0.81}$ & $38.42_{\pm 2.83}$ & $-14.87_{\pm 3.15}$ & $50.51_{\pm 2.90}$ & $-8.46_{\pm 4.18}$ \\
TreeLoRA & $46.07_{\pm 3.14}$ & $-8.50_{\pm 2.24}$ & $39.30_{\pm 2.44}$ & $-12.69_{\pm 2.21}$ & $47.58_{\pm 1.35}$ & $-11.52_{\pm 2.12}$ \\
MoE-CL & $47.31_{\pm 0.77}$ & $-6.32_{\pm 0.88}$ & $46.61_{\pm 0.48}$ & $-4.55_{\pm 1.17}$ & $53.08_{\pm 0.77}$ & $-4.69_{\pm 1.49}$ \\
\midrule
SETA & $\mathbf{48.39}_{\pm 1.10}$ & $\mathbf{-1.96}_{\pm 0.57}$ & $\mathbf{47.90}_{\pm 1.07}$ & $\mathbf{-2.20}_{\pm 0.69}$ & $\mathbf{54.73}_{\pm 0.14}$ & $\mathbf{-1.28}_{\pm 0.45}$ \\
% SETA ($\beta$=25\%) & $\mathbf{49.43}_{\pm\ 1.21}$ & $\mathbf{-1.54}_{\pm\ 0.89}$ & $\mathbf{52.86}_{\pm\ 2.16}$ & $\mathbf{0.67}_{\pm\ 0.1.2}$ & $\mathbf{55.36}_{\pm\ 2.82}$ & $-4.00_{\pm 1.81}$ \\
\bottomrule
\end{tabular}
\end{table*}
\begin{figure*}[t]
\centering
\includegraphics[width=0.95\textwidth]{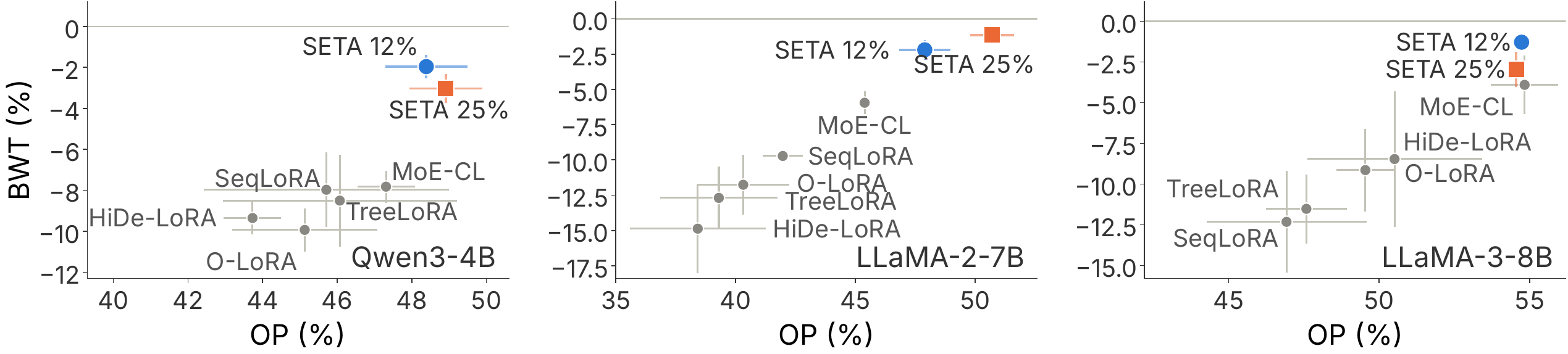}
\caption{Final performance (OP) against backward transfer (BWT) on TRACE-8 for three backbones. Points are means over
seeds 100 and 42, and error bars show one standard deviation on each axis. The horizontal line marks zero forgetting,
so methods toward the upper right both learn more and forget less. SETA is shown at block budgets of 12\% and 25\%.}
\label{fig:stability_plasticity}
\end{figure*}
\subsection{Adaptive Gating Expansion}
\label{sec:gating_expansion}
\textbf{Logit Invariance via Weight Inheritance:}
Let $W_g^{(t-1)}$ be the gate row of the active expert at task $t-1$, with routing logit
$z = \mathbf{x}^\top W_g^{(t-1)} + b_g^{(t-1)}$ for a token $\mathbf{x}$ (Figure~\ref{fig:seta_architecture}d). When an
expert is split, both successors inherit the parent row in full, $W_g^{(\text{s})} \leftarrow W_g^{(t-1)}$ and
$W_g^{(\text{u})} \leftarrow W_g^{(t-1)}$. Under independent sigmoid gating this preserves each successor's
activation exactly rather than halving it, with a perturbation of 2\% of the row norm added to $W_g^{(\text{s})}$
to break symmetry. An expert with no ancestor has its gate bias opened so that it trains at an activation comparable
to that of the single expert at $T_1$. All gate rows, including those of earlier experts, stay trainable while task $t$ is trained; inheritance and the opened bias only keep routing unchanged at the moment of the split and prevent
cold-start errors.

\subsection{Task-Agnostic Inference with Independent Sigmoid Gating}
\label{sec:sigmoid_gating}
Routing is content-based, without task identifiers: each adapted projection has its own gating network, a single
linear layer $W_g \in \mathbb{R}^{N_l \times d}$ with bias $\mathbf{b}_g$, producing one logit for each of the
$N_l$ experts that hold blocks in that layer (Figure~\ref{fig:seta_architecture}e). All experts are evaluated
together:
\begin{equation}
\label{eq:task_agnostic_pred}
\begin{split}
    \mathbf{y} = W_0 \mathbf{x} + \sum_{k \in \mathcal{E}_{\text{s}}} \sigma_k(\mathbf{x} W_g^\top + \mathbf{b}_g) \Delta W_k \mathbf{x} \\
    + \sum_{k \in \mathcal{E}_{\text{u}}} \sigma_k(\mathbf{x} W_g^\top + \mathbf{b}_g) \Delta W_k \mathbf{x},
\end{split}
\end{equation}
where $\sigma(\cdot)$ is the element-wise sigmoid, so each expert is gated independently,
$g_i(\mathbf{x}) = \sigma(\mathbf{x}^\top W_g^{(i)} + b_i)$, rather than normalized across experts. A softmax gate
would enforce $\sum_i g_i = 1$, which fails here for two reasons. First, expert deltas are summed,
and a token may need several experts at full strength, such as two experts
holding copies of the same block; a softmax caps their combined weight at one. Second, normalization couples the rows: $\partial g_i / \partial z_t < 0$ for $i \neq t$, so every
step raising the current task's logit suppresses all prior experts. Sigmoid removes both, giving
$\partial g_i / \partial z_t = 0$ for $i \neq t$.
\section{Experiments}
\label{sec:experiments}
\subsection{Experimental Setup}
\textbf{Benchmarks:}
We evaluate on TRACE~\cite{wang2023trace} and the O-LoRA benchmark~\cite{wang2023orthogonal}, including its fifteen-task long-sequence setting. TRACE-8 trains C-STANCE, FOMC, MeetingBank, Py150, ScienceQA, NumGLUE-cm, NumGLUE-ds and 20Minuten in this order, covering all four TRACE categories. The O-LoRA benchmark has two settings: three orders of four text-classification tasks (AG News, Amazon, DBpedia,
Yahoo), and three orders of the fifteen tasks of \citet{razdaibiedina2023progressive}, which add Yelp, IMDB, four GLUE
tasks and five SuperGLUE tasks. We run LLaMA-2-7B, LLaMA-3-8B and Qwen3-4B on TRACE-8, and Qwen3-4B on the O-LoRA benchmark. All methods use the plain instruction prompt of Appendix~\ref{app:baseline}. Unless stated otherwise, all experiments use 5,000 training samples and 500 test samples per task.

\textbf{Baselines:}
SeqLoRA trains one adapter on the tasks in sequence; O-LoRA~\cite{wang2023orthogonal} learns each task in a
subspace orthogonal to earlier ones; HiDe-LoRA~\cite{wang2024hide} and TreeLoRA~\cite{qian2025treelora} combine
task-specific and shared LoRA components; MoE-CL~\cite{kang2025moecl} pairs a LoRA expert per task with a shared
LoRA expert trained against a task discriminator. FIX prompts the frozen backbone with in-context examples.

\textbf{Metrics:}
For a stream of $T$ tasks, let $M_{i,j}$ denote the score on task $j$ after the model has been trained on task $i$.
We report three metrics. Learning Accuracy, $\mathrm{LA}=\frac{1}{T}\sum_{j=1}^{T}M_{j,j}$, is the average score on
each task right after it is learned and measures plasticity. Overall Performance,
$\mathrm{OP}=\frac{1}{T}\sum_{j=1}^{T}M_{T,j}$, is the average score of the final model on all tasks. Backward
Transfer, $\mathrm{BWT}=\frac{1}{T-1}\sum_{j=1}^{T-1}(M_{T,j}-M_{j,j})$, is the average change on earlier tasks by
the end of the stream; a negative value indicates forgetting. Higher is better for all three. We score
classification and reasoning tasks by accuracy, Py150 by edit similarity, MeetingBank by ROUGE-L and 20Minuten by
SARI.

\textbf{Block budget:} SETA adapts only the V projection. Its block budget $\beta$ is the fraction of V-projection blocks each task may
select, so it plays the same role as the rank in LoRA and sets how many parameters a task can train. We use
$\beta$=12\% throughout unless stated otherwise, and also test $\beta$=25\%, which raises the trainable parameters
per task by a factor of 1.4 to 1.8.
\begin{table*}[htbp]
\centering
\caption{Comparison of continual learning methods on the O-LoRA benchmark with Qwen3-4B. O1--O3 are orders of the
four standard text-classification tasks and O4--O6 orders of the fifteen-task long sequence, following
\citet{wang2023orthogonal} and \citet{razdaibiedina2023progressive}. We report OP and BWT (\%) after the final task as mean$_{\pm\text{std}}$ over two
seeds. \textbf{Bold} marks the best mean for each metric in each row.}
\label{tab:olora_summary}
\small
\setlength{\tabcolsep}{2.5pt}
\resizebox{\textwidth}{!}{%
\begin{tabular}{lcccccccccccc}
\toprule
& \multicolumn{2}{c}{SeqLoRA} & \multicolumn{2}{c}{O-LoRA} & \multicolumn{2}{c}{HiDe-LoRA} & \multicolumn{2}{c}{TreeLoRA} & \multicolumn{2}{c}{MoE-CL} & \multicolumn{2}{c}{SETA} \\
\cmidrule(lr){2-3}\cmidrule(lr){4-5}\cmidrule(lr){6-7}\cmidrule(lr){8-9}\cmidrule(lr){10-11}\cmidrule(lr){12-13}
Order & OP & BWT & OP & BWT & OP & BWT & OP & BWT & OP & BWT & OP & BWT \\
\midrule
O1 & $70.00_{\pm 0.00}$ & $-6.67_{\pm 1.41}$ & $70.00_{\pm 1.06}$ & $-7.00_{\pm 0.00}$ & $69.50_{\pm 0.35}$ & $-5.83_{\pm 0.24}$ & $69.50_{\pm 0.35}$ & $-7.50_{\pm 1.18}$ & $70.00_{\pm 1.77}$ & $-7.33_{\pm 2.83}$ & $\mathbf{75.50}_{\pm 1.06}$ & $\mathbf{-2.00}_{\pm 0.94}$ \\
O2 & $71.38_{\pm 1.24}$ & $-6.67_{\pm 0.47}$ & $71.00_{\pm 0.71}$ & $-6.67_{\pm 2.36}$ & $69.75_{\pm 0.00}$ & $-6.00_{\pm 0.47}$ & $71.50_{\pm 1.06}$ & $-6.00_{\pm 1.89}$ & $68.12_{\pm 1.59}$ & $-9.33_{\pm 1.41}$ & $\mathbf{77.00}_{\pm 0.35}$ & $\mathbf{-1.33}_{\pm 0.47}$ \\
O3 & $74.00_{\pm 0.35}$ & $-4.00_{\pm 0.47}$ & $73.38_{\pm 0.18}$ & $-5.00_{\pm 0.00}$ & $72.50_{\pm 0.35}$ & $-4.17_{\pm 1.65}$ & $73.50_{\pm 0.35}$ & $-4.33_{\pm 0.00}$ & $73.88_{\pm 0.18}$ & $-2.33_{\pm 0.00}$ & $\mathbf{74.00}_{\pm 1.06}$ & $\mathbf{-1.67}_{\pm 0.47}$ \\
\midrule
O4 & $76.15_{\pm 0.20}$ & $-4.06_{\pm 0.26}$ & $75.87_{\pm 1.23}$ & $-4.35_{\pm 1.27}$ & $74.42_{\pm 2.19}$ & $-5.33_{\pm 2.05}$ & $76.31_{\pm 0.02}$ & $-3.84_{\pm 0.02}$ & $76.33_{\pm 0.28}$ & $\mathbf{-3.42}_{\pm 0.15}$ & $\mathbf{77.14}_{\pm 2.06}$ & $-4.16_{\pm 1.86}$ \\
O5 & $79.17_{\pm 0.33}$ & $-1.49_{\pm 0.57}$ & $79.44_{\pm 0.26}$ & $-1.02_{\pm 0.45}$ & $79.36_{\pm 0.16}$ & $\mathbf{-0.41}_{\pm 0.03}$ & $78.65_{\pm 0.05}$ & $-2.00_{\pm 0.38}$ & $78.01_{\pm 0.75}$ & $-1.94_{\pm 0.24}$ & $\mathbf{80.06}_{\pm 2.89}$ & $-0.80_{\pm 0.42}$ \\
O6 & $79.00_{\pm 0.35}$ & $-0.92_{\pm 0.31}$ & $78.98_{\pm 0.39}$ & $-0.78_{\pm 0.49}$ & $78.50_{\pm 0.22}$ & $-0.87_{\pm 0.12}$ & $78.85_{\pm 0.58}$ & $-0.99_{\pm 0.70}$ & $78.89_{\pm 1.05}$ & $-1.18_{\pm 1.04}$ & $\mathbf{79.89}_{\pm 0.67}$ & $\mathbf{-0.58}_{\pm 0.25}$ \\
\bottomrule
\end{tabular}}
\end{table*}
\begin{figure*}[t]
  \centering
\includegraphics[width=0.90\textwidth]{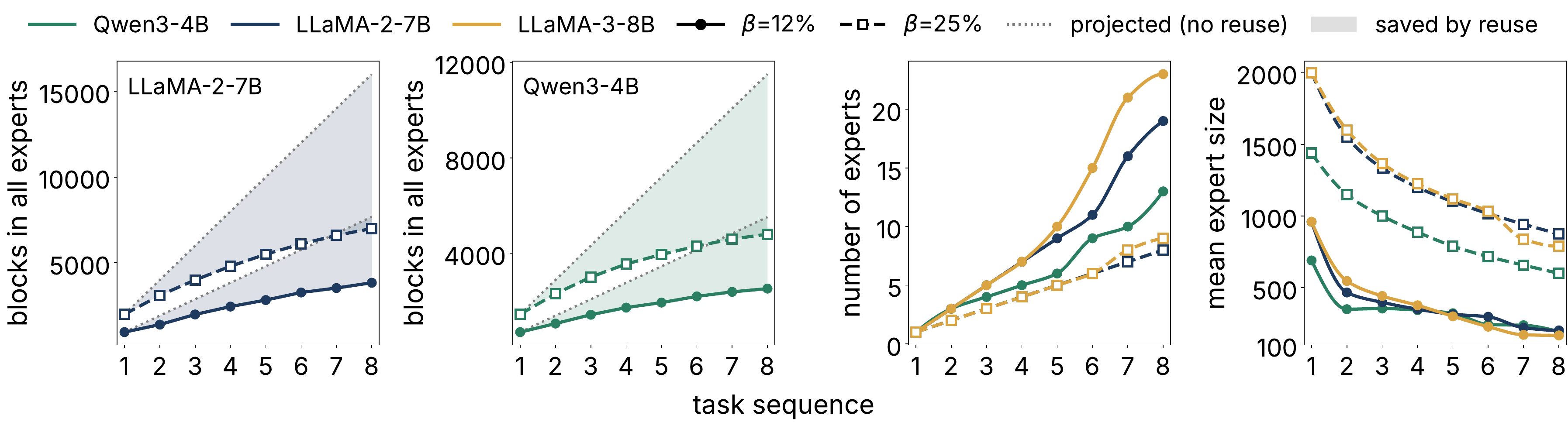}
\caption{Expert structure of SETA along the TRACE-8 task sequence at block budgets of 12\% (solid) and 25\% (dashed).
(a) LLaMA-2-7B and (b) Qwen3-4B: distinct V-projection block positions used by any expert, against the count if every
task took fresh blocks (dotted); the shaded gap is the number of positions reused by later tasks (reuse saves
positions, not stored parameters; Appendix~\ref{app:growth}). (c) Number of experts and (d) mean expert size in
blocks for all three backbones; at 25\% each task adds exactly one expert.}
\label{fig:reuse_panels}
\end{figure*}

\subsection{Results on TRACE-8}
\label{sec:exp_main}
\textbf{Retention:}
SETA's main advantage is that it forgets little (Table~\ref{tab:all_models_summary}). Its BWT is between $-2.2$ and
$-1.3$ on all three backbones, while SeqLoRA, O-LoRA, HiDe-LoRA and TreeLoRA lose 8 to 15 points and MoE-CL, the strongest baseline, 4.6 to 6.3. In SETA a new task trains its own unique expert, the shared experts it joins and the gate rows; earlier unique experts stay frozen, and elastic anchoring keeps the shared experts near their earlier values. Earlier tasks can therefore change only through shared experts and routing. The LoRA baselines keep updating weights that earlier tasks use; orthogonal subspaces and task-specific branches reduce this interference but do not remove it. SETA has the best OP on every backbone, by 1.3 points on LLaMA-2-7B, 1.7 on LLaMA-3-8B and 1.1 on Qwen3-4B over MoE-CL. On Qwen3-4B, forgetting cancels what most of the baselines learn, and the frozen FIX model beats SeqLoRA, O-LoRA, HiDe-LoRA and TreeLoRA; only MoE-CL and SETA score above it.

\subsection{Task Order and Sequence Length}
\label{sec:exp_robust}

\textbf{Short sequences:}
On three orders of the same four tasks (Table~\ref{tab:olora_summary}, O1--O3), SETA has the best or tied-best OP
on every order. The tasks use different label sets, so baselines that keep updating shared weights overwrite earlier
label mappings and forget 2.3 to 9.3 points. SETA keeps each mapping mostly in its own expert and forgets at most two.

\textbf{Long sequences:}
The 15-task streams (O4--O6) repeat task types, so later tasks partly rehearse earlier ones and every method forgets
less, especially on O5 and O6. Isolation then matters less: SETA still has the best mean OP on all three orders, but on O4 and O5 the margin lies within its own seed spread (std 2.06 and 2.89).
\subsection{Subspace Efficiency}
\label{sec:exp_eff}

\textbf{Training cost per task:}
At $\beta$=12\%, SETA trains 0.2 to 1\% of the backbone per task (Appendix~\ref{app:growth}): the task's own new
expert, the shared experts it joins and the small gating networks. Earlier unique experts are never updated again, and the warm-up that scores
the blocks stops on its own after a short phase.

\textbf{Effect of the block budget:}
Raising $\beta$ to 25\% gives each task 1.4 to 1.8 times more trainable parameters and raises LA on all three
backbones, but it turns into higher OP only on LLaMA-2-7B (+2.8 OP, BWT from $-2.2$ to $-1.2$;
Figure~\ref{fig:stability_plasticity}). On Qwen3-4B (+0.5) and LLaMA-3-8B ($-0.2$) the OP change is within one
standard deviation, because forgetting grows (BWT $-2.0 \to -3.0$ and $-1.3 \to -3.0$), mostly on NumGLUE and
MeetingBank for Qwen3-4B and on ScienceQA and Py150 for LLaMA-3-8B. 

\textbf{Subspace reuse:}
Different tasks put their largest gradients on many of the same V-projection blocks
(Appendix~\ref{app:phase0:overlap}), so the union of task subspaces grows more slowly than the number of tasks
(Figure~\ref{fig:reuse_panels}a--b). After eight tasks Qwen3-4B uses less than half the blocks that disjoint subspaces would need. To keep later tasks from crowding onto the same blocks, selection skips any block already held by two tasks.
\begin{table}[htbp]
\centering
\caption{Routing rules at inference on TRACE-8 with Qwen3-4B (seed 100, OP \%). Each final checkpoint is evaluated
with SETA's own gates, oracle routing by task identity, or only the top-$k$ gates. Keeping only the top-$k$ gates
lowers OP at both budgets, so predictions draw on several experts at once.}
\label{tab:routing_avgs}
\small
\begin{tabular}{lcc}
\toprule
Routing & $\beta = 12\%$ & $\beta = 25\%$ \\
\midrule
Task-Agnostic & 49.16 & 49.60 \\
Oracle  & 49.21 & 51.09 \\
Top-1   & 32.53 & 34.20 \\
Top-2   & 27.69 & 37.83 \\
Top-3   & 32.65 & 40.00 \\
\bottomrule
\end{tabular}
\end{table}

\begin{table}[htbp]
\centering
\caption{Effect of the direction test on TRACE-8 with Qwen3-4B (reduced protocol, seed 100). Strict uses $Z$=3,
$\tau$=+0.02; normal $Z$=2, $\tau$=0; relaxed $Z$=1.5, $\tau$=$-$0.02; the default $Z$=3, $\tau$=0. Passed and Shared
count blocks over tasks 2--8 that pass the test and that stay shared after the size rule; Undone is the share of
passed blocks the size rule removes, and Cos$<$.02 the share with gradient cosine below 0.02.}
\label{tab:dir_main}
\setlength{\tabcolsep}{2.5pt}
\resizebox{\columnwidth}{!}{%
\begin{tabular}{lccccccc}
\toprule
Test & $\varphi$ & Passed & Shared & Undone (\%) & Cos$<$.02 (\%) & OP & BWT \\
\midrule
Strict  & 0.006 & 1446 & 468  & 67.6 & 0.0$^{*}$ & 34.69 & $-$6.29 \\
Normal  & 0.006 & 3226 & 1074 & 66.7 & 31.2      & 34.74 & $-$5.56 \\
Relaxed & 0.006 & 3851 & 1182 & 69.3 & 37.7      & 34.65 & $-$6.66 \\
\midrule
Default & 0.012 & 2358 & 179  & 92.4 & 33.0      & 37.15 & $-$4.95 \\
\bottomrule
\end{tabular}}
\end{table}
\begin{table}[htbp]
\centering
\small
\setlength{\tabcolsep}{3.5pt}
\caption{Sensitivity to the anchoring strength $\lambda_{\text{base}}$ on TRACE-8 with Qwen3-4B,
under the reduced protocol of Table~\ref{tab:abl_reduced}. All runs use the sign test and differ only in
$\lambda_{\text{base}}$; $\Delta$OP is the change from the default, and $\lambda_{\text{base}}$=0 turns anchoring
off. The default gives both the highest OP and the least forgetting.}
\label{tab:abl_knobs_lam}
\begin{tabular}{lcccc}
\toprule
$\lambda_{\text{base}}$ & LA & OP & BWT & $\Delta$OP \\
\midrule
0 (no anchoring) & 40.56 & 34.34 & $-$7.11 & $-$2.81 \\
0.05 & 40.60 & 34.96 & $-$6.44 & $-$2.19 \\
0.1 (default) & 41.48 & 37.15 & $-$4.95 &  \\
0.2 & 42.24 & 36.79 & $-$6.22 & $-$0.36 \\
0.5 & 42.35 & 35.31 & $-$8.05 & $-$1.84 \\
\bottomrule
\end{tabular}
\end{table}

\subsection{Contribution of Each Component}
\label{sec:exp_abl}
\begin{table}[t]
\centering
\caption{Ablations of SETA on TRACE-8 with Qwen3-4B under the full protocol of
Table~\ref{tab:all_models_summary} (5,000 training samples per task, seed 100),
changing the adapted projection, the shared experts.
$\Delta$OP is the change from the first row.}
\label{tab:abl_main}
\small
\begin{tabular}{lcccc}
\toprule
Variant & LA $\uparrow$ & OP $\uparrow$ & BWT $\uparrow$ & $\Delta$OP \\
\midrule
SETA ($\beta$=12\%, V) & 50.53 & 49.16 & $-$1.56 &  \\
Q projection & 40.93 & 34.90 & $-$6.89 & $-$14.26 \\
MLP projections & 42.44 & 35.82 & $-$7.56 & $-$13.34 \\
No shared experts & 52.35 & 51.70 & $-$0.85 & +2.54 \\
\bottomrule
\end{tabular}
\end{table}

\begin{table}[htbp]
\centering
\caption{Component ablations of SETA on TRACE-8 with Qwen3-4B under a reduced
protocol (500 training samples per task, fixed warm-up, seed 100). Each row changes one setting of the full system,
which here freezes the gate rows of earlier experts. $\Delta$OP is relative to the first row. Scores are not
comparable with Table~\ref{tab:abl_main}.}
\label{tab:abl_reduced}
\setlength{\tabcolsep}{3pt}
\resizebox{\columnwidth}{!}{%
\begin{tabular}{lcccc}
\toprule
Variant & LA $\uparrow$ & OP $\uparrow$ & BWT $\uparrow$ & $\Delta$OP \\
\midrule
SETA, all components & 41.48 & 37.15 & $-$4.95 &  \\
Sign test off & 40.96 & 34.77 & $-$7.07 & $-$2.38 \\
No direction test (positional split) & 39.46 & 35.37 & $-$4.67 & $-$1.78 \\
No projection of own copies & 40.97 & 37.51 & $-$3.96 & +0.36 \\
Earlier gate rows trainable & 40.93 & 35.89 & $-$5.76 & $-$1.26 \\
No elastic anchoring ($\lambda_{\text{base}}$=0) & 40.56 & 34.34 & $-$7.11 & $-$2.81 \\
\bottomrule
\end{tabular}}
\end{table}
\textbf{Choice of projection:}
Adapting V rather than Q is the most important design choice: with Q, OP drops by 14.26 points
(Table~\ref{tab:abl_main}). Adapting Q both learns less (LA 40.93 against 50.53) and forgets more (BWT $-6.89$
against $-1.56$); right after training, MeetingBank reaches only 8.5 ROUGE-L with Q against 23.4 with V. About 90--94\%
of high-magnitude gradients fall in V (Figure~\ref{fig:gradient_analysis}), which we attribute to softmax saturation
shrinking the gradients that reach Q.

\textbf{Shared experts, split rule and anchoring:}
Under the reduced protocol (Table~\ref{tab:abl_reduced}), removing elastic anchoring lowers OP by 2.81, the sign
test by 2.38 and the direction test by 1.78, while the projection of a task's own copies is not needed (+0.36). Under the full protocol, removing shared experts raises OP by 2.54
(Table~\ref{tab:abl_main}), so with only 108 shared blocks over the sequence this run does not show that sharing
helps. All differences come from one seed and are below the 4.7-point gap between the two SeqLoRA seeds.

\textbf{Task-agnostic routing:}
Keeping only the top-$k$ gates lowers OP by 16.5 to 21.5 points at $\beta$=12\% and by 9.6 to 15.4 at
$\beta$=25\% (Table~\ref{tab:routing_avgs}), because tasks also draw on experts formed by other tasks. Without task
identifiers, SETA's gates come within 0.05 points of oracle routing at $\beta$=12\% and fall 1.5 points short at
$\beta$=25\%.

\textbf{Direction test versus size floor:}
We compare a strict test ($Z$=3, $\tau$=+0.02), a normal test ($Z$=2, $\tau$=0) and a relaxed test ($Z$=1.5,
$\tau$=$-$0.02) at $\varphi$=0.006 with the default ($Z$=3, $\tau$=0, $\varphi$=0.012).
The test finds clear agreement only between closely related tasks: at the second task, where two classification tasks
meet, fewer than 1\% of the passing blocks have a gradient cosine below 0.02, but over the whole stream a third of the
blocks the default test passes do (Appendix~\ref{app:growth}). The relaxed test passes 2.7 times and shares 2.5 times
as many blocks as the strict test with OP unchanged (34.65 to 34.74).
\textbf{Robustness to split and anchoring settings:} SETA is robust to most settings. Loosening the direction test more than doubles the blocks that pass, yet OP stays
within 0.1 points because the size floor undoes most of the extra sharing. Only a
size floor of $\varphi$=0.04 breaks the method ($-5.74$ OP): it exceeds a typical layer's selection, so most layers
are not adapted. Anchoring works best at the default $\lambda_{\text{base}}$=0.1:
weaker anchoring lets shared experts drift, and the strongest setting learns most, but also forgets most.
\section{Conclusion}
We presented SETA, a framework for mitigating catastrophic forgetting in large language models under continual learning. By separating task-specific and shared knowledge within a dynamically evolving Mixture-of-Experts architecture, SETA isolates conflicting updates while preserving reusable representations through adaptive elastic anchoring. Experimental results show that this structural decomposition enables task-agnostic inference and achieves competitive or superior performance relative to state-of-the-art continual learning baselines.
\section*{Limitations}
Our study has three main limitations. First, we compare SETA only with parameter-efficient baselines.
Continual learning methods that update all model weights need considerably more GPU memory than our setup provides, so we leave a comparison
with them to future work. Second, our experiments use models with 4B to 8B parameters, and it remains to be seen
whether the same behavior holds for larger models. Third, we evaluate SETA only on language models; applying it to
vision and multimodal models is a natural next step.
\section{Acknowledgment}
This research was supported by the Argonne Given Internship Program and U.S. Department of Energy, Office of Science, Advanced Scientific Computing Research, under contract number DE-AC02-06CH11357. This research used resources of the Argonne Leadership Computing Facility at Argonne National Laboratory, which is supported by the Office of Science of the U.S. Department of Energy, Office of Science, under contract number DE-AC02-06CH11357. We appreciate the support from the ISU Dean’s Emerging Faculty Leaders Award. We also thank the National Center for Supercomputing Applications for providing Delta GPUs through allocation CIS240604 from the Advanced Cyberinfrastructure Coordination Ecosystem: Services \& Support (ACCESS) program.
\bibliography{custom}

@article{brown2020language,
  title={Language models are few-shot learners},
  author={Brown, Tom and Mann, Benjamin and Ryder, Nick and Subbiah, Melanie and Kaplan, Jared D and Dhariwal, Prafulla and Neelakantan, Arvind and Shyam, Pranav and Sastry, Girish and Askell, Amanda and others},
  journal={Advances in neural information processing systems},
  volume={33},
  pages={1877--1901},
  year={2020}
}

@article{raffel2020exploring,
  title={Exploring the limits of transfer learning with a unified text-to-text transformer},
  author={Raffel, Colin and Shazeer, Noam and Roberts, Adam and Lee, Katherine and Narang, Sharan and Matena, Michael and Zhou, Yanqi and Li, Wei and Liu, Peter J},
  journal={Journal of machine learning research},
  volume={21},
  number={140},
  pages={1--67},
  year={2020}
}

@article{parisi2019continual,
  title={Continual lifelong learning with neural networks: A review},
  author={Parisi, German I and Kemker, Ronald and Part, Jose L and Kanan, Christopher and Wermter, Stefan},
  journal={Neural networks},
  volume={113},
  pages={54--71},
  year={2019},
  publisher={Elsevier}
}

@article{li2017learning,
  title={Learning without forgetting},
  author={Li, Zhizhong and Hoiem, Derek},
  journal={IEEE transactions on pattern analysis and machine intelligence},
  volume={40},
  number={12},
  pages={2935--2947},
  year={2017},
  publisher={IEEE}
}

@article{chaudhry2019tiny,
  title={On tiny episodic memories in continual learning},
  author={Chaudhry, Arslan and Rohrbach, Marcus and Elhoseiny, Mohamed and Ajanthan, Thalaiyasingam and Dokania, Puneet K and Torr, Philip HS and Ranzato, Marc'Aurelio},
  journal={arXiv preprint arXiv:1902.10486},
  year={2019}
}

@article{kirkpatrick2017overcoming,
  title={Overcoming catastrophic forgetting in neural networks},
  author={Kirkpatrick, James and Pascanu, Razvan and Rabinowitz, Neil and Veness, Joel and Desjardins, Guillaume and Rusu, Andrei A and Milan, Kieran and Quan, John and Ramalho, Tiago and Grabska-Barwinska, Agnieszka and others},
  journal={Proceedings of the national academy of sciences},
  volume={114},
  number={13},
  pages={3521--3526},
  year={2017},
  publisher={National Academy of Sciences}
}

@article{rusu2016progressive,
  title={Progressive neural networks},
  author={Rusu, Andrei A and Rabinowitz, Neil C and Desjardins, Guillaume and Soyer, Hubert and Kirkpatrick, James and Kavukcuoglu, Koray and Pascanu, Razvan and Hadsell, Raia},
  journal={arXiv preprint arXiv:1606.04671},
  year={2016}
}

@article{han2015learning,
  title={Learning both weights and connections for efficient neural network},
  author={Han, Song and Pool, Jeff and Tran, John and Dally, William},
  journal={Advances in neural information processing systems},
  volume={28},
  year={2015}
}

@inproceedings{evci2020rigging,
  title={Rigging the lottery: Making all tickets winners},
  author={Evci, Utku and Gale, Trevor and Menick, Jacob and Castro, Pablo Samuel and Elsen, Erich},
  booktitle={International conference on machine learning},
  pages={2943--2952},
  year={2020},
  organization={PMLR}
}

@article{wu2024continual,
  title={Continual learning for large language models: A survey},
  author={Wu, Tongtong and Luo, Linhao and Li, Yuan-Fang and Pan, Shirui and Vu, Thuy-Trang and Haffari, Gholamreza},
  journal={arXiv preprint arXiv:2402.01364},
  year={2024}
}

@article{hu2022lora,
  title={Lora: Low-rank adaptation of large language models.},
  author={Hu, Edward J and Shen, Yelong and Wallis, Phillip and Allen-Zhu, Zeyuan and Li, Yuanzhi and Wang, Shean and Wang, Liang and Chen, Weizhu and others},
  journal={Iclr},
  volume={1},
  number={2},
  pages={3},
  year={2022}
}

@inproceedings{ren2024analyzing,
  title={Analyzing and reducing catastrophic forgetting in parameter efficient tuning},
  author={Li, Xinlong and Ren, Weijieying and Qin, Wei and Wang, Lei and Zhao, Tianxiang and Hong, Richang},
  booktitle={ICASSP 2025-2025 IEEE International Conference on Acoustics, Speech and Signal Processing (ICASSP)},
  pages={1--5},
  year={2025},
  organization={IEEE}
}

@inproceedings{lopez2017gradient,
  title={Gradient episodic memory for continual learning},
  author={Lopez-Paz, David and Ranzato, Marc'Aurelio},
  booktitle={Advances in Neural Information Processing Systems},
  volume={30},
  year={2017}
}

@article{razdaibiedina2023progressive,
  title={Progressive prompts: Continual learning for language models},
  author={Razdaibiedina, Anastasia and Mao, Yuning and Hou, Rui and Khabsa, Madian and Lewis, Mike and Almahairi, Amjad},
  journal={arXiv preprint arXiv:2301.12314},
  year={2023}
}

@inproceedings{li2021prefix,
  title={Prefix-tuning: Optimizing continuous prompts for generation},
  author={Li, Xiang Lisa and Liang, Percy},
  booktitle={Proceedings of the 59th Annual Meeting of the Association for Computational Linguistics and the 11th International Joint Conference on Natural Language Processing (Volume 1: Long Papers)},
  pages={4582--4597},
  year={2021}
}

@incollection{mccloskey1989catastrophic,
  title={Catastrophic interference in connectionist networks: The sequential learning problem},
  author={McCloskey, Michael and Cohen, Neal J},
  booktitle={Psychology of learning and motivation},
  volume={24},
  pages={109--165},
  year={1989},
  publisher={Elsevier}
}

@inproceedings{rebuffi2017icarl,
  title={icarl: Incremental classifier and representation learning},
  author={Rebuffi, Sylvestre-Alvise and Kolesnikov, Alexander and Sperl, Georg and Lampert, Christoph H},
  booktitle={Proceedings of the IEEE conference on Computer Vision and Pattern Recognition},
  pages={2001--2010},
  year={2017}
}

@inproceedings{li2024ccpeft,
  title={Customizable Combination of Parameter-Efficient Modules for Multi-Task Learning},
  author={Li, Han and Xu, Yuchen and others},
  booktitle={Proceedings of EMNLP},
  year={2024}
}

@inproceedings{wang2023orthogonal,
  title={Orthogonal subspace learning for language model continual learning},
  author={Wang, Xiao and Chen, Tianze and Ge, Qiming and Xia, Han and Bao, Rong and Zheng, Rui and Zhang, Qi and Gui, Tao and Huang, Xuan-Jing},
  booktitle={Findings of the Association for Computational Linguistics: EMNLP 2023},
  pages={10658--10671},
  year={2023}
}

@inproceedings{han2025slim,
  title={Slim: Let llm learn more and forget less with soft lora and identity mixture},
  author={Han, Jiayi and Du, Liang and Du, Hongwei and Zhou, Xiangguo and Wu, Yiwen and Zhang, Yuanfang and Zheng, Weibo and Han, Donghong},
  booktitle={Proceedings of the 2025 Conference of the Nations of the Americas Chapter of the Association for Computational Linguistics: Human Language Technologies (Volume 1: Long Papers)},
  pages={4792--4804},
  year={2025}
}

@inproceedings{houlsby2019parameter,
  title={Parameter-efficient transfer learning for NLP},
  author={Houlsby, Neil and Giurgiu, Andrei and Jastrzebski, Stanislaw and Morrone, Bruna and De Laroussilhe, Quentin and Gesmundo, Andrea and Attariyan, Mona and Gelly, Sylvain},
  booktitle={International conference on machine learning},
  pages={2790--2799},
  year={2019},
  organization={PMLR}
}

@inproceedings{zaken2022bitfit,
  title={Bitfit: Simple parameter-efficient fine-tuning for transformer-based masked language-models},
  author={Zaken, Elad Ben and Goldberg, Yoav and Ravfogel, Shauli},
  booktitle={Proceedings of the 60th Annual Meeting of the Association for Computational Linguistics (Volume 2: Short Papers)},
  pages={1--9},
  year={2022}
}

@inproceedings{lester2021power,
  title={The power of scale for parameter-efficient prompt tuning},
  author={Lester, Brian and Al-Rfou, Rami and Constant, Noah},
  booktitle={Proceedings of the 2021 conference on empirical methods in natural language processing},
  pages={3045--3059},
  year={2021}
}

@inproceedings{lialin2024relora,
  title={Relora: High-rank training through low-rank updates},
  author={Lialin, Vladislav and Muckatira, Sherin and Shivagunde, Namrata and Rumshisky, Anna},
  booktitle={International Conference on Learning Representations},
  volume={2024},
  pages={49405--49421},
  year={2024}
}

@article{zhou2024rankadaptor,
  title={Rankadaptor: Hierarchical dynamic low-rank adaptation for structural pruned llms},
  author={Zhou, Changhai and Han, Shijie and Zhang, Shiyang and Weng, Shichao and Liu, Zekai and Jin, Cheng},
  journal={arXiv preprint arXiv:2406.15734},
  year={2024},
  publisher={Jun}
}

@inproceedings{yang2024adazeta,
  title={Adazeta: Adaptive zeroth-order tensor-train adaption for memory-efficient large language models fine-tuning},
  author={Yang, Yifan and Zhen, Kai and Banijamali, Ershad and Mouchtaris, Athanasios and Zhang, Zheng},
  booktitle={Proceedings of the 2024 Conference on Empirical Methods in Natural Language Processing},
  pages={977--995},
  year={2024}
}

@inproceedings{gao2024adaptive,
  title={Adaptive rank selections for low-rank approximation of language models},
  author={Gao, Shangqian and Hua, Ting and Hsu, Yen-Chang and Shen, Yilin and Jin, Hongxia},
  booktitle={Proceedings of the 2024 Conference of the North American Chapter of the Association for Computational Linguistics: Human Language Technologies (Volume 1: Long Papers)},
  pages={227--241},
  year={2024}
}

@article{li2024mixlora,
  title={Mixlora: Enhancing large language models fine-tuning with lora-based mixture of experts},
  author={Li, Dengchun and Ma, Yingzi and Wang, Naizheng and Ye, Zhengmao and Cheng, Zhiyuan and Tang, Yinghao and Zhang, Yan and Duan, Lei and Zuo, Jie and Yang, Cal and others},
  journal={arXiv preprint arXiv:2404.15159},
  year={2024}
}

@inproceedings{huang2024mixture,
  title={Mixture of lora experts},
  author={Huang, Shaohan and Wei, Furu and others},
  booktitle={International Conference on Learning Representations},
  volume={2024},
  pages={47302--47318},
  year={2024}
}

@article{liu2024adamole,
  title={Adamole: Fine-tuning large language models with adaptive mixture of low-rank adaptation experts},
  author={Liu, Zefang and Luo, Jiahua},
  journal={arXiv preprint arXiv:2405.00361},
  year={2024}
}

@article{munoz2025low,
  title={Low-rank adapters meet neural architecture search for llm compression},
  author={Mu{\~n}oz, J Pablo and Yuan, Jinjie and Jain, Nilesh},
  journal={arXiv preprint arXiv:2501.16372},
  year={2025}
}

@inproceedings{he2025smt,
  title={Smt: Fine-tuning large language models with sparse matrices},
  author={He, Haoze and Li, Juncheng and Jiang, Xuan and Miller, Heather},
  booktitle={International Conference on Learning Representations},
  volume={2025},
  pages={32369--32393},
  year={2025}
}

@article{shazeer2017outrageously,
  title={Outrageously large neural networks: The sparsely-gated mixture-of-experts layer},
  author={Shazeer, Noam and Mirhoseini, Azalia and Maziarz, Krzysztof and Davis, Andy and Le, Quoc and Hinton, Geoffrey and Dean, Jeff},
  journal={arXiv preprint arXiv:1701.06538},
  year={2017}
}

@article{lepikhin2020gshard,
  title={Gshard: Scaling giant models with conditional computation and automatic sharding},
  author={Lepikhin, Dmitry and Lee, HyoukJoong and Xu, Yuanzhong and Chen, Dehao and Firat, Orhan and Huang, Yanping and Krikun, Maxim and Shazeer, Noam and Chen, Zhifeng},
  journal={arXiv preprint arXiv:2006.16668},
  year={2020}
}

@article{fedus2022switch,
  title={Switch transformers: Scaling to trillion parameter models with simple and efficient sparsity},
  author={Fedus, William and Zoph, Barret and Shazeer, Noam},
  journal={Journal of Machine Learning Research},
  volume={23},
  number={120},
  pages={1--39},
  year={2022}
}

@inproceedings{lewis2021base,
  title={Base layers: Simplifying training of large, sparse models},
  author={Lewis, Mike and Bhosale, Shruti and Dettmers, Tim and Goyal, Naman and Zettlemoyer, Luke},
  booktitle={International Conference on Machine Learning},
  pages={6265--6274},
  year={2021},
  organization={PMLR}
}

@article{roller2021hash,
  title={Hash layers for large sparse models},
  author={Roller, Stephen and Sukhbaatar, Sainbayar and Weston, Jason and others},
  journal={advances in neural information processing systems},
  volume={34},
  pages={17555--17566},
  year={2021}
}

@article{zhou2022mixture,
  title={Mixture-of-experts with expert choice routing},
  author={Zhou, Yanqi and Lei, Tao and Liu, Hanxiao and Du, Nan and Huang, Yanping and Zhao, Vincent and Dai, Andrew M and Le, Quoc V and Laudon, James and others},
  journal={Advances in Neural Information Processing Systems},
  volume={35},
  pages={7103--7114},
  year={2022}
}

@article{lu2022learn,
  title={Learn to explain: Multimodal reasoning via thought chains for science question answering},
  author={Lu, Pan and Mishra, Swaroop and Xia, Tanglin and Qiu, Liang and Chang, Kai-Wei and Zhu, Song-Chun and Tafjord, Oyvind and Clark, Peter and Kalyan, Ashwin},
  journal={Advances in neural information processing systems},
  volume={35},
  pages={2507--2521},
  year={2022}
}

@inproceedings{shah2023trillion,
  title={Trillion dollar words: A new financial dataset, task \& market analysis},
  author={Shah, Agam and Paturi, Suvan and Chava, Sudheer},
  booktitle={Proceedings of the 61st Annual Meeting of the Association for Computational Linguistics (Volume 1: Long Papers)},
  pages={6664--6679},
  year={2023}
}

@inproceedings{hu2023meetingbank,
  title={Meetingbank: A benchmark dataset for meeting summarization},
  author={Hu, Yebowen and Ganter, Timothy and Deilamsalehy, Hanieh and Dernoncourt, Franck and Foroosh, Hassan and Liu, Fei},
  booktitle={Proceedings of the 61st Annual Meeting of the Association for Computational Linguistics (Volume 1: Long Papers)},
  pages={16409--16423},
  year={2023}
}

@inproceedings{zhao2023c,
  title={C-STANCE: A large dataset for Chinese zero-shot stance detection},
  author={Zhao, Chenye and Li, Yingjie and Caragea, Cornelia},
  booktitle={Proceedings of the 61st Annual Meeting of the Association for Computational Linguistics (Volume 1: Long Papers)},
  pages={13369--13385},
  year={2023}
}

@inproceedings{kew202320,
  title={20 Minuten: A Multi-task News Summarisation Dataset for German},
  author={Kew, Tannon and Kostrzewa, Marek and Ebling, Sarah},
  booktitle={Proceedings of the 8th edition of the Swiss Text Analytics Conference},
  pages={1--13},
  year={2023}
}

@inproceedings{mishra2022numglue,
  title={NumGLUE: A suite of fundamental yet challenging mathematical reasoning tasks},
  author={Mishra, Swaroop and Mitra, Arindam and Varshney, Neeraj and Sachdeva, Bhavdeep and Clark, Peter and Baral, Chitta and Kalyan, Ashwin},
  booktitle={Proceedings of the 60th Annual Meeting of the Association for Computational Linguistics (Volume 1: Long Papers)},
  pages={3505--3523},
  year={2022}
}

@article{wang2023trace,
  title={TRACE: A comprehensive benchmark for continual learning in large language models},
  author={Wang, Xiao and Zhang, Yuansen and Chen, Tianze and Gao, Songyang and Jin, Senjie and Yang, Xianjun and Xi, Zhiheng and Zheng, Rui and Zou, Yicheng and Gui, Tao and others},
  journal={arXiv preprint arXiv:2310.06762},
  year={2023}
}

@article{wang2024hide,
  title={Hide-pet: continual learning via hierarchical decomposition of parameter-efficient tuning},
  author={Wang, Liyuan and Xie, Jingyi and Zhang, Xingxing and Su, Hang and Zhu, Jun},
  journal={IEEE Transactions on Pattern Analysis and Machine Intelligence},
  volume={47},
  number={8},
  pages={6687--6702},
  year={2025},
  publisher={IEEE}
}

@article{qian2025treelora,
  title={Treelora: Efficient continual learning via layer-wise loras guided by a hierarchical gradient-similarity tree},
  author={Qian, Yu-Yang and Xu, Yuan-Ze and Zhang, Zhen-Yu and Zhao, Peng and Zhou, Zhi-Hua},
  journal={arXiv preprint arXiv:2506.10355},
  year={2025}
}

@article{zhang2015character,
  title={Character-level convolutional networks for text classification},
  author={Zhang, Xiang and Zhao, Junbo and LeCun, Yann},
  journal={Advances in neural information processing systems},
  volume={28},
  year={2015}
}

@article{lu2021codexglue,
  title={Codexglue: A machine learning benchmark dataset for code understanding and generation},
  author={Lu, Shuai and Guo, Daya and Ren, Shuo and Huang, Junjie and Svyatkovskiy, Alexey and Blanco, Ambrosio and Clement, Colin and Drain, Dawn and Jiang, Daxin and Tang, Duyu and others},
  journal={arXiv preprint arXiv:2102.04664},
  year={2021}
}

@inproceedings{wang2018glue,
  title={GLUE: A multi-task benchmark and analysis platform for natural language understanding},
  author={Wang, Alex and Singh, Amanpreet and Michael, Julian and Hill, Felix and Levy, Omer and Bowman, Samuel R},
  booktitle={Proceedings of the 2018 EMNLP workshop BlackboxNLP: Analyzing and interpreting neural networks for NLP},
  pages={353--355},
  year={2018}
}

@article{wang2019superglue,
  title={Superglue: A stickier benchmark for general-purpose language understanding systems},
  author={Wang, Alex and Pruksachatkun, Yada and Nangia, Nikita and Singh, Amanpreet and Michael, Julian and Hill, Felix and Levy, Omer and Bowman, Samuel},
  journal={Advances in neural information processing systems},
  volume={32},
  year={2019}
}

@inproceedings{maas2011learning,
  title={Learning word vectors for sentiment analysis},
  author={Maas, Andrew and Daly, Raymond E and Pham, Peter T and Huang, Dan and Ng, Andrew Y and Potts, Christopher},
  booktitle={Proceedings of the 49th annual meeting of the association for computational linguistics: Human language technologies},
  pages={142--150},
  year={2011}
}

@inproceedings{kang2025moecl,
  title={Self-evolving llms via continual instruction tuning},
  author={Kang, Jiazheng and Huang, Le and Hou, Cheng and Zhao, Zhe and Yan, ZhenXiang and Bai, Ting},
  booktitle={Proceedings of the ACM Web Conference 2026},
  pages={7724--7732},
  year={2026}
}

@article{hossen2026craft,
  title   = {{CRAFT}: Forgetting-Aware Intervention-Based Adaptation for Continual Learning},
  author  = {Hossen, Md Anwar and Siddika, Fatema and Munoz, Juan Pablo and Roosta, Tanya and Jannesari, Ali},
  journal = {arXiv preprint arXiv:2605.05732},
  year    = {2026}
}

@misc{anonymous2026fedseam,
  title  = {{FedSEAM}: Federated Continual Learning of {LLMs} via Subspace Expert Agreement Merging},
  author = {Anonymous},
  year   = {2026},
  note   = {Under review}
}
\appendix
\section{Extended Related Work} \label{app:Exrelated}
\label{appendix:related_works}
\subsection{Continual Learning}
Continual learning (CL) seeks to train models on sequential tasks without catastrophic forgetting ~\citep{mccloskey1989catastrophic}, where newly acquired knowledge overwrites previously learned representations. 
Classical CL methods include regularization-based approaches such as LwF~\citep{li2017learning}, rehearsal-based replay~\citep{rebuffi2017icarl}, and architectural expansion~\citep{rusu2016progressive}, though these often require large memory buffers or suffer from scalability issues. Recent advances extend CL to Transformer-based and large language models (LLMs) using parameter-efficient tuning (PEFT) modules. 
SLiM~\citep{han2025slim} introduces a soft mixture of LoRA and identity mappings, dynamically weighting learned and frozen components to enhance retention while preserving efficiency. 
Meanwhile, CC-PEFT~\citep{li2024ccpeft} generalizes modular parameter-efficient tuning by combining multiple PEFT adapters in a customizable manner for multi-task and continual scenarios. 

\subsection{Mixture-of-Experts (MoE)} 
Traditional dense models process every input token with all parameters, making computation scale linearly with model size. 
Mixture-of-Experts (MoE) improves efficiency by activating only a subset of parameters, first shown effective in LSTMs \citep{shazeer2017outrageously} and later integrated into Transformers \citep{lepikhin2020gshard, fedus2022switch}. 
Subsequent work addressed load imbalance with routing strategies \citep{lewis2021base, roller2021hash, zhou2022mixture}. 

\section*{Appendix}
\section{Sensitivity of the Split Rule}
\label{app:split_sensitivity}
The two tables in this appendix study the split rule of Section~\ref{sec:split_on_share} on TRACE-8 with Qwen3-4B.
Both use the reduced protocol of Table~\ref{tab:abl_reduced} (500 training samples per task, seed 100), so their
scores are compared only with each other, not with Table~\ref{tab:all_models_summary}.
Table~\ref{tab:dir-ablation} varies how strict the direction test is and reports, for each setting, how many
overlapping blocks pass the test, how many stay shared after the size rule, and the resulting LA, OP and BWT. The
strict, normal and relaxed tests differ only in $Z$ and $\tau$: the relaxed test shares 2.5 times as many blocks as
the strict one, yet OP changes by less than 0.1.
Table~\ref{tab:abl_knobs_rule} changes one split-rule setting at a time (the agreement bar $Z$, the size floor
$\varphi$ and the kept fraction share$_q$) in an earlier round of runs; only a size floor larger than a typical
layer's selection ($\varphi$=0.04) clearly hurts.

\begin{table*}[htbp]
\caption{Effect of the direction test's strictness on Qwen3-4B (TRACE-8, reduced protocol with the sign test on).
$Z$ is the agreement bar, $\tau$ the minimum gradient cosine for sharing and $\varphi$ the minimum expert size as a
fraction of the block budget. Split structure counts the blocks that pass the direction test, those kept as shared
after the size rule, the share of passing blocks undone by that rule, the share of passing blocks with a gradient
cosine below 0.02, and the experts after the last task. A looser test passes and shares more blocks, but the size
rule undoes most of them and performance barely changes among the three arms at $\varphi$=0.006. $^{*}$Zero by
construction.}
\label{tab:dir-ablation}
\centering
\small
\setlength{\tabcolsep}{4.5pt}
\begin{tabular}{lccc rrrrr ccc}
\toprule
 & \multicolumn{3}{c}{Setting} & \multicolumn{5}{c}{Split structure (tasks 2--8)} & \multicolumn{3}{c}{Performance} \\
\cmidrule(lr){2-4}\cmidrule(lr){5-9}\cmidrule(lr){10-12}
Arm & $Z$ & $\tau$ & $\varphi$ & Passed & Shared & Undone (\%) & cos$<$0.02 (\%) & Experts & LA & OP & BWT \\
\midrule
Strict      & 3.0 & $+0.02$ & 0.006 & 1446 & 468  & 67.6 & 0.0$^{*}$ & 26 & 40.19 & 34.69 & $-6.29$ \\
Normal      & 2.0 & $0.00$  & 0.006 & 3226 & 1074 & 66.7 & 31.2      & 41 & 39.60 & 34.74 & $-5.56$ \\
Relaxed     & 1.5 & $-0.02$ & 0.006 & 3851 & 1182 & 69.3 & 37.7      & 43 & 40.47 & 34.65 & $-6.66$ \\
\midrule
Default        & 3.0 & $0.00$  & 0.012 & 2358 & 179  & 92.4 & 33.0      & 17 & 41.48 & 37.15 & $-4.95$ \\
\bottomrule
\end{tabular}
\end{table*}

\section{Detailed Results}
\label{app:results}
Table~\ref{tab:domain_specific_results} gives the per-task scores on TRACE-8, including SETA at $\beta$=25\%, and
Figure~\ref{fig:stability_plasticity} places every method by its final performance (OP) and forgetting (BWT).

\paragraph{Stability and plasticity:}
In Figure~\ref{fig:stability_plasticity}, SETA sits in the upper right of every panel and the baselines in the
lower left. The four LoRA baselines of Table~\ref{tab:domain_specific_results} learn each task slightly better (up to 3.7 points of LA), but they lose 8 to 15 points on
earlier tasks, which costs them more than they gained; MoE-CL, which has no per-task scores in that table, loses 4.6 to 6.3. A larger budget moves SETA right and up on LLaMA-2-7B, while
on Qwen3-4B and LLaMA-3-8B it moves down by about 1 to 1.7 BWT points with OP nearly unchanged.

\begin{table*}[t]
\centering
\caption{Per-task scores (\%) on TRACE-8 in training order. For each trained method, the first row gives the score
on each task after the last task (its mean is OP) and the second row the score right after that task was learned
(its mean is LA), so comparing the two rows shows where each method forgets. All entries are
mean$_{\pm\text{std}}$ over two seeds (FIX runs on the LLaMA backbones have no per-task scores).}
\label{tab:domain_specific_results}
\setlength{\tabcolsep}{3pt}
\resizebox{\textwidth}{!}{%
\begin{tabular}{lcccccccccc}
\toprule
Method & C-STANCE & FOMC & MeetingB. & Py150 & SciQA & NG-cm & NG-ds & 20Min. & OP $\uparrow$ & BWT $\uparrow$ \\
\midrule
\multicolumn{11}{l}{\textit{LLaMA-2-7B}} \\
FIX &  &  &  &  &  &  &  &  & $30.29_{\pm 1.86}$ & -- \\
SeqLoRA & $32.8_{\pm 2.5}$ & $46.9_{\pm 3.6}$ & $35.9_{\pm 0.5}$ & $42.0_{\pm 3.2}$ & $54.5_{\pm 7.2}$ & $37.0_{\pm 3.5}$ & $46.2_{\pm 0.4}$ & $40.4_{\pm 0.8}$ & $41.97_{\pm 0.84}$ & $-9.72_{\pm 0.13}$ \\
\quad \textit{after own task} & $51.2_{\pm 1.1}$ & $65.7_{\pm 0.6}$ & $39.2_{\pm 0.3}$ & $60.9_{\pm 0.1}$ & $58.7_{\pm 1.6}$ & $37.0_{\pm 3.5}$ & $50.6_{\pm 0.7}$ & $40.4_{\pm 0.8}$ & & \\
O-LoRA & $22.3_{\pm 17.4}$ & $43.6_{\pm 0.7}$ & $36.8_{\pm 0.0}$ & $42.2_{\pm 11.0}$ & $52.6_{\pm 5.4}$ & $35.8_{\pm 5.2}$ & $48.5_{\pm 0.7}$ & $40.7_{\pm 0.2}$ & $40.32_{\pm 1.93}$ & $-11.76_{\pm 2.11}$ \\
\quad \textit{after own task} & $51.2_{\pm 0.8}$ & $65.9_{\pm 1.1}$ & $39.4_{\pm 0.2}$ & $60.9_{\pm 0.1}$ & $59.6_{\pm 2.0}$ & $38.3_{\pm 3.5}$ & $48.8_{\pm 0.2}$ & $40.7_{\pm 0.2}$ & & \\
HiDe-LoRA & $13.1_{\pm 8.3}$ & $47.8_{\pm 4.6}$ & $36.1_{\pm 0.1}$ & $32.0_{\pm 16.2}$ & $53.7_{\pm 3.5}$ & $37.0_{\pm 10.5}$ & $47.4_{\pm 4.4}$ & $40.3_{\pm 0.9}$ & $38.42_{\pm 2.83}$ & $-14.87_{\pm 3.15}$ \\
\quad \textit{after own task} & $50.4_{\pm 0.3}$ & $66.4_{\pm 0.4}$ & $39.2_{\pm 0.7}$ & $61.0_{\pm 0.0}$ & $59.0_{\pm 0.8}$ & $43.8_{\pm 0.9}$ & $51.2_{\pm 0.7}$ & $40.3_{\pm 0.9}$ & & \\
TreeLoRA & $16.5_{\pm 18.2}$ & $45.8_{\pm 3.4}$ & $36.7_{\pm 2.3}$ & $42.5_{\pm 4.3}$ & $48.4_{\pm 2.5}$ & $36.4_{\pm 2.6}$ & $48.3_{\pm 0.0}$ & $39.9_{\pm 0.3}$ & $39.30_{\pm 2.44}$ & $-12.69_{\pm 2.21}$ \\
\quad \textit{after own task} & $51.1_{\pm 0.1}$ & $66.9_{\pm 0.3}$ & $39.2_{\pm 0.4}$ & $60.9_{\pm 0.1}$ & $57.9_{\pm 3.0}$ & $37.0_{\pm 1.7}$ & $50.3_{\pm 0.2}$ & $39.9_{\pm 0.3}$ & & \\
SETA ($\beta$=12\%) & $48.9_{\pm 1.6}$ & $63.9_{\pm 1.1}$ & $35.7_{\pm 0.7}$ & $46.0_{\pm 4.2}$ & $56.1_{\pm 0.2}$ & $36.4_{\pm 1.0}$ & $56.5_{\pm 0.6}$ & $40.0_{\pm 0.1}$ & $47.90_{\pm 1.07}$ & $-2.20_{\pm 0.69}$ \\
\quad \textit{after own task} & $49.6_{\pm 1.1}$ & $70.4_{\pm 1.3}$ & $40.8_{\pm 1.3}$ & $43.0_{\pm 0.2}$ & $65.1_{\pm 2.1}$ & $36.5_{\pm 0.9}$ & $53.1_{\pm 1.1}$ & $40.0_{\pm 0.1}$ & & \\
SETA ($\beta$=25\%) & $55.5_{\pm 0.7}$ & $71.1_{\pm 1.1}$ & $38.1_{\pm 1.1}$ & $51.5_{\pm 1.7}$ & $68.0_{\pm 1.4}$ & $29.6_{\pm 0.2}$ & $52.0_{\pm 4.3}$ & $39.9_{\pm 0.1}$ & $50.72_{\pm 0.91}$ & $-1.15_{\pm 0.05}$ \\
\quad \textit{after own task} & $40.0_{\pm 13.8}$ & $75.3_{\pm 1.5}$ & $42.8_{\pm 0.5}$ & $45.1_{\pm 1.2}$ & $76.8_{\pm 1.9}$ & $39.8_{\pm 0.1}$ & $54.0_{\pm 1.1}$ & $39.9_{\pm 0.1}$ & & \\
\midrule
\multicolumn{11}{l}{\textit{LLaMA-3-8B}} \\
FIX &  &  &  &  &  &  &  &  & $36.50_{\pm 0.11}$ & -- \\
SeqLoRA & $35.6_{\pm 2.5}$ & $44.9_{\pm 17.8}$ & $28.8_{\pm 0.7}$ & $50.8_{\pm 3.4}$ & $70.1_{\pm 0.1}$ & $52.5_{\pm 2.6}$ & $53.8_{\pm 0.4}$ & $38.8_{\pm 1.2}$ & $46.92_{\pm 2.66}$ & $-12.32_{\pm 3.12}$ \\
\quad \textit{after own task} & $51.1_{\pm 1.0}$ & $70.3_{\pm 0.4}$ & $39.9_{\pm 0.4}$ & $64.8_{\pm 0.7}$ & $76.8_{\pm 0.6}$ & $64.2_{\pm 3.5}$ & $55.7_{\pm 1.7}$ & $38.8_{\pm 1.2}$ & & \\
O-LoRA & $43.6_{\pm 0.8}$ & $48.6_{\pm 7.7}$ & $28.5_{\pm 0.4}$ & $50.9_{\pm 1.4}$ & $74.5_{\pm 3.3}$ & $56.8_{\pm 0.0}$ & $54.0_{\pm 1.1}$ & $39.5_{\pm 0.3}$ & $49.54_{\pm 0.96}$ & $-9.14_{\pm 2.55}$ \\
\quad \textit{after own task} & $50.9_{\pm 1.3}$ & $69.7_{\pm 0.4}$ & $39.3_{\pm 0.1}$ & $64.4_{\pm 0.6}$ & $77.0_{\pm 0.3}$ & $64.2_{\pm 7.0}$ & $55.4_{\pm 2.2}$ & $39.5_{\pm 0.3}$ & & \\
HiDe-LoRA & $41.1_{\pm 1.3}$ & $48.1_{\pm 12.7}$ & $31.6_{\pm 1.8}$ & $53.5_{\pm 0.7}$ & $74.9_{\pm 2.1}$ & $58.6_{\pm 6.1}$ & $57.5_{\pm 0.9}$ & $38.6_{\pm 0.2}$ & $50.51_{\pm 2.90}$ & $-8.46_{\pm 4.18}$ \\
\quad \textit{after own task} & $51.0_{\pm 0.3}$ & $68.9_{\pm 0.1}$ & $38.7_{\pm 0.5}$ & $64.6_{\pm 0.0}$ & $77.9_{\pm 1.0}$ & $64.8_{\pm 6.1}$ & $58.8_{\pm 0.9}$ & $38.6_{\pm 0.2}$ & & \\
TreeLoRA & $36.9_{\pm 8.1}$ & $46.5_{\pm 5.0}$ & $30.3_{\pm 1.2}$ & $47.6_{\pm 0.2}$ & $74.1_{\pm 0.7}$ & $51.9_{\pm 0.0}$ & $54.0_{\pm 0.7}$ & $39.5_{\pm 0.1}$ & $47.58_{\pm 1.35}$ & $-11.52_{\pm 2.12}$ \\
\quad \textit{after own task} & $51.0_{\pm 1.1}$ & $70.1_{\pm 0.1}$ & $38.5_{\pm 0.4}$ & $64.6_{\pm 0.3}$ & $76.9_{\pm 1.3}$ & $64.2_{\pm 3.5}$ & $56.6_{\pm 2.6}$ & $39.5_{\pm 0.1}$ & & \\
SETA ($\beta$=12\%) & $61.5_{\pm 0.7}$ & $72.5_{\pm 0.7}$ & $43.2_{\pm 0.2}$ & $52.2_{\pm 0.3}$ & $48.5_{\pm 0.7}$ & $64.6_{\pm 0.7}$ & $57.5_{\pm 0.7}$ & $37.8_{\pm 0.1}$ & $54.73_{\pm 0.14}$ & $-1.28_{\pm 0.45}$ \\
\quad \textit{after own task} & $51.0_{\pm 0.6}$ & $69.1_{\pm 0.1}$ & $44.5_{\pm 0.4}$ & $50.2_{\pm 0.3}$ & $64.1_{\pm 1.0}$ & $71.6_{\pm 0.2}$ & $58.6_{\pm 0.6}$ & $37.8_{\pm 0.1}$ & & \\
SETA ($\beta$=25\%) & $65.5_{\pm 0.7}$ & $76.5_{\pm 0.7}$ & $41.4_{\pm 0.1}$ & $39.4_{\pm 1.6}$ & $49.0_{\pm 0.0}$ & $62.9_{\pm 0.1}$ & $62.5_{\pm 0.7}$ & $39.2_{\pm 0.1}$ & $54.55_{\pm 0.11}$ & $-2.95_{\pm 1.07}$ \\
\quad \textit{after own task} & $52.0_{\pm 1.2}$ & $70.8_{\pm 1.4}$ & $44.0_{\pm 1.3}$ & $51.1_{\pm 0.3}$ & $72.8_{\pm 1.3}$ & $67.1_{\pm 1.3}$ & $60.1_{\pm 1.6}$ & $39.2_{\pm 0.1}$ & & \\
\midrule
\multicolumn{11}{l}{\textit{Qwen3-4B}} \\
FIX & $59.8_{\pm 1.1}$ & $55.1_{\pm 3.8}$ & $13.3_{\pm 0.3}$ & $4.1_{\pm 0.1}$ & $87.8_{\pm 0.8}$ & $63.6_{\pm 2.6}$ & $50.5_{\pm 0.9}$ & $38.3_{\pm 0.3}$ & $46.55_{\pm 0.72}$ & -- \\
SeqLoRA & $30.0_{\pm 16.4}$ & $45.1_{\pm 3.6}$ & $26.7_{\pm 1.8}$ & $33.2_{\pm 2.7}$ & $81.7_{\pm 0.1}$ & $53.7_{\pm 13.1}$ & $56.5_{\pm 0.7}$ & $38.9_{\pm 0.0}$ & $45.71_{\pm 3.29}$ & $-7.97_{\pm 1.83}$ \\
\quad \textit{after own task} & $57.1_{\pm 0.1}$ & $64.9_{\pm 1.1}$ & $30.5_{\pm 0.1}$ & $43.9_{\pm 0.3}$ & $83.9_{\pm 0.1}$ & $44.4_{\pm 15.7}$ & $57.8_{\pm 1.3}$ & $38.9_{\pm 0.0}$ & & \\
O-LoRA & $26.0_{\pm 11.9}$ & $42.8_{\pm 4.4}$ & $26.8_{\pm 0.1}$ & $33.6_{\pm 5.2}$ & $80.6_{\pm 0.8}$ & $56.2_{\pm 2.6}$ & $56.0_{\pm 1.3}$ & $39.0_{\pm 0.5}$ & $45.13_{\pm 1.94}$ & $-9.93_{\pm 1.06}$ \\
\quad \textit{after own task} & $57.2_{\pm 0.0}$ & $64.9_{\pm 1.1}$ & $30.2_{\pm 0.0}$ & $44.0_{\pm 0.2}$ & $83.7_{\pm 0.1}$ & $54.9_{\pm 9.6}$ & $56.6_{\pm 0.4}$ & $39.0_{\pm 0.5}$ & & \\
HiDe-LoRA & $24.7_{\pm 20.5}$ & $41.3_{\pm 9.1}$ & $26.5_{\pm 0.6}$ & $36.2_{\pm 2.8}$ & $83.7_{\pm 0.1}$ & $43.8_{\pm 4.4}$ & $54.2_{\pm 0.9}$ & $39.4_{\pm 0.4}$ & $43.73_{\pm 0.77}$ & $-9.35_{\pm 0.81}$ \\
\quad \textit{after own task} & $56.9_{\pm 0.1}$ & $66.3_{\pm 0.6}$ & $30.4_{\pm 1.0}$ & $44.0_{\pm 0.1}$ & $84.4_{\pm 0.0}$ & $38.3_{\pm 1.7}$ & $55.5_{\pm 1.1}$ & $39.4_{\pm 0.4}$ & & \\
TreeLoRA & $29.6_{\pm 10.5}$ & $44.1_{\pm 7.6}$ & $26.3_{\pm 0.0}$ & $33.0_{\pm 1.8}$ & $81.5_{\pm 2.1}$ & $57.4_{\pm 6.1}$ & $57.5_{\pm 0.9}$ & $39.1_{\pm 0.3}$ & $46.07_{\pm 3.14}$ & $-8.50_{\pm 2.24}$ \\
\quad \textit{after own task} & $57.3_{\pm 0.1}$ & $65.7_{\pm 1.4}$ & $30.4_{\pm 0.0}$ & $43.8_{\pm 0.4}$ & $83.4_{\pm 0.3}$ & $51.9_{\pm 10.5}$ & $56.5_{\pm 0.7}$ & $39.1_{\pm 0.3}$ & & \\
SETA ($\beta$=12\%) & $66.0_{\pm 2.8}$ & $64.0_{\pm 1.4}$ & $27.2_{\pm 1.0}$ & $48.2_{\pm 1.8}$ & $79.0_{\pm 1.4}$ & $41.3_{\pm 1.2}$ & $23.9_{\pm 1.2}$ & $37.5_{\pm 0.4}$ & $48.39_{\pm 1.10}$ & $-1.96_{\pm 0.57}$ \\
\quad \textit{after own task} & $56.9_{\pm 0.1}$ & $65.5_{\pm 1.5}$ & $23.0_{\pm 0.4}$ & $48.0_{\pm 1.1}$ & $75.0_{\pm 0.0}$ & $47.9_{\pm 1.1}$ & $47.0_{\pm 0.3}$ & $37.5_{\pm 0.4}$ & & \\
SETA ($\beta$=25\%) & $63.0_{\pm 0.0}$ & $65.0_{\pm 1.4}$ & $23.9_{\pm 1.2}$ & $48.0_{\pm 0.4}$ & $72.0_{\pm 1.4}$ & $47.5_{\pm 1.9}$ & $32.4_{\pm 2.0}$ & $39.5_{\pm 0.4}$ & $48.91_{\pm 0.98}$ & $-3.03_{\pm 0.71}$ \\
\quad \textit{after own task} & $56.9_{\pm 0.3}$ & $65.9_{\pm 0.3}$ & $31.8_{\pm 1.6}$ & $50.2_{\pm 0.5}$ & $59.5_{\pm 1.2}$ & $59.6_{\pm 0.2}$ & $49.1_{\pm 0.6}$ & $39.5_{\pm 0.4}$ & & \\
\bottomrule
\end{tabular}}
\end{table*}

\paragraph{Per-task retention:}
The clearest contrast is on the first task, C-STANCE, which every later task can overwrite. The baselines fall from
about 51 right after training to 13.1--32.8 on LLaMA-2-7B and 35.6--43.6 on LLaMA-3-8B, and from about 57 to
24.7--30.0 on Qwen3-4B, with seed deviations up to 20 points. SETA's C-STANCE score instead rises by 6.1 to 13.5
points on LLaMA-3-8B and Qwen3-4B and stays level on LLaMA-2-7B at $\beta$=12\% (the large gain at $\beta$=25\%
comes from one low seed right after training). SETA's largest losses are NumGLUE-ds on Qwen3-4B at both budgets
(47.0 to 23.9 and 49.1 to 32.4), ScienceQA on LLaMA-3-8B at both budgets (64.1 to 48.5 and 72.8 to 49.0), and
ScienceQA (65.1 to 56.1) and NumGLUE-cm (39.8 to 29.6) on LLaMA-2-7B at the two budgets.
\section{SETA Builds and Shares Experts}
\label{app:experts}
This section follows SETA's experts through TRACE-8: how many blocks each task selects and keeps, how the expert set grows on each backbone (Appendix~\ref{app:growth}), which
overlapping blocks the direction-aware Split-on-Share rule turns into shared experts,
how the minimum expert size limits sharing (Appendix~\ref{app:min_size}), and how large the learned updates of
unique and shared experts are.

\subsection{Expert Growth Across Backbones}
\label{app:growth}
\begin{table*}[t]
\centering
\small
\caption{Expert structure of SETA after each task on TRACE-8 for three backbones at block budgets at $\beta=12\%$, 
read from the split reports and training logs. Experts counts all experts after the task's split; shared blocks
formed counts the blocks moved into new shared experts; layers kept counts the layers that retain the task's blocks
after the minimum-size rule; trainable parameters are the experts updated for the task, as a share of the
backbone. At the larger budget no shared expert forms, so each task adds one isolated expert (Figure~\ref{fig:reuse_panels}c).}
\label{tab:expert_growth}
\setlength{\tabcolsep}{4pt}
\begin{tabular}{lcccccccc}
\toprule
& C-ST & FOMC & MeetB & Py150 & SciQA & NG-cm & NG-ds & 20Min \\
\midrule
\multicolumn{9}{l}{\textit{$\beta$=12\%}} \\
Qwen3-4B, experts & 1 & 3 & 4 & 5 & 6 & 9 & 10 & 13 \\
\quad shared blocks formed & 0 & 55 & 0 & 0 & 0 & 22 & 0 & 31 \\
\quad layers kept (of 36) & 34 & 33 & 34 & 34 & 34 & 30 & 33 & 31 \\
\quad trainable params (\%) & 0.28 & 0.28 & 0.28 & 0.28 & 0.28 & 0.27 & 0.28 & 0.28 \\
LLaMA-2-7B, experts & 1 & 3 & 5 & 7 & 9 & 11 & 16 & 19 \\
\quad shared blocks formed & 0 & 176 & 13 & 16 & 17 & 42 & 80 & 36 \\
\quad layers kept (of 32) & 32 & 31 & 31 & 31 & 32 & 32 & 32 & 32 \\
\quad trainable params (\%) & 1.01 & 1.00 & 0.99 & 0.98 & 0.98 & 0.98 & 0.98 & 0.97 \\
LLaMA-3-8B, experts & 1 & 3 & 5 & 7 & 10 & 15 & 21 & 23 \\
\quad shared blocks formed & 0 & 82 & 32 & 14 & 25 & 83 & 130 & 12 \\
\quad layers kept (of 32) & 32 & 30 & 32 & 30 & 30 & 30 & 30 & 31 \\
\quad trainable params (\%) & 0.20 & 0.20 & 0.20 & 0.20 & 0.20 & 0.21 & 0.21 & 0.21 \\
\bottomrule
\end{tabular}
\end{table*}

\begin{figure*}[t]
  \centering
  \includegraphics[width=0.92\textwidth]{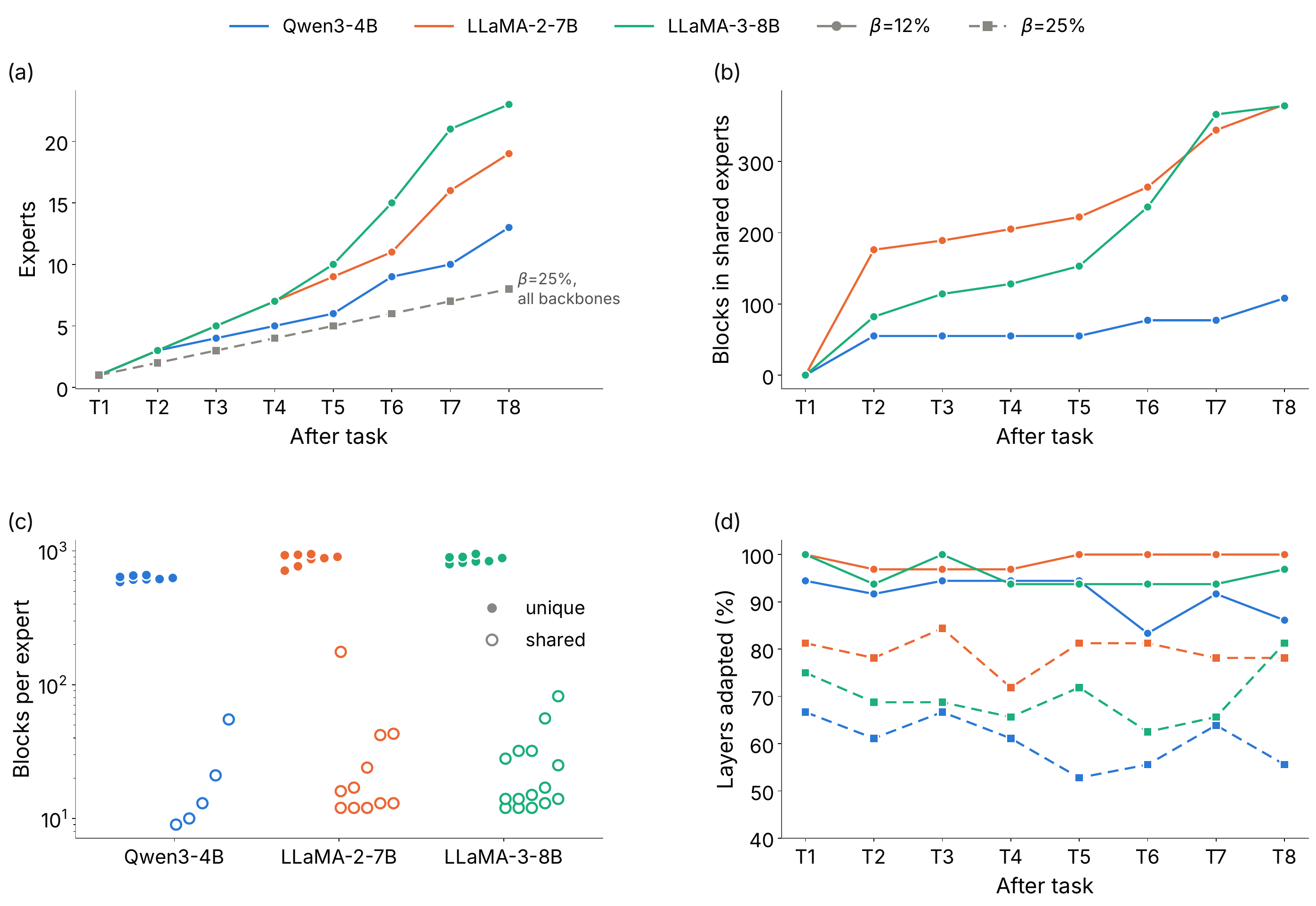}
  \caption{How SETA's experts grow on TRACE-8. (a) Number of experts after each task's split; at the
  larger budget every task adds exactly one unique expert. (b) Blocks held by shared experts at $\beta$=12\% (none
  form at 25\%). (c) Size of every expert after the last task at $\beta$=12\% (log scale), unique (filled) against
  shared (hollow). (d) Share of the backbone's layers that keep the task's blocks after the minimum-size rule.
  Shared experts stay small, so most of each task's update remains in its own frozen expert.}
  \label{fig:expert_growth}
\end{figure*}

Table~\ref{tab:expert_growth} and Figure~\ref{fig:expert_growth} extend the LLaMA-2-7B analysis to all three
backbones. At $\beta$=12\%, every task adds one unique expert and most splits add a few small shared experts (on
Qwen3-4B only three of the seven splits do), so the expert set grows to 13 experts on Qwen3-4B, 19 on LLaMA-2-7B and 23 on LLaMA-3-8B after eight tasks. The shared
experts stay small: after the last task they hold 108, 380 and 378 blocks in total, between 9 and 176 blocks each,
while a unique expert holds 586 to 660 blocks on Qwen3-4B, 712 to 947 on LLaMA-2-7B and 796 to 950 on LLaMA-3-8B.
Sharing therefore covers a small part of each task's update, and each task keeps most of it in its own frozen
expert. This also bounds the memory that sharing saves, since a shared block replaces one copy per task
that holds it: after eight tasks the experts store 5,115 blocks on Qwen3-4B against 5,223 without sharing, 7,327
against 7,707 on LLaMA-2-7B and 7,297 against 7,675 on LLaMA-3-8B, a saving of 2 to 5\%. At $\beta$=25\%, no shared expert forms on any backbone, so each task adds exactly one expert and the
model is a set of isolated experts (Figure~\ref{fig:reuse_panels}c; Appendix~\ref{app:min_size}).

The trained share of the backbone depends on the block size. With $256\times256$ blocks, LLaMA-2-7B trains 0.97 to
1.01\% of its parameters per task at $\beta$=12\%, while Qwen3-4B and LLaMA-3-8B, with $128\times128$ blocks, train
0.27 to 0.28\% and 0.20 to 0.21\%. At $\beta$=25\% the share rises by a factor of 1.4 to 1.8 rather than 2, since
the minimum-size rule removes more layers (Figure~\ref{fig:expert_growth}d).

The subspace test rejects 51\%,
49\% and 68\% of the overlaps on Qwen3-4B, LLaMA-2-7B and LLaMA-3-8B, and its rejection rate is lowest at the
second task, when the new task meets a single earlier expert. Of the overlaps
that pass the direction test, the minimum-size rule returns most: only 108 of 1,795 passing overlaps become shared
blocks on Qwen3-4B, 380 of 3,531 on LLaMA-2-7B and 378 of 1,925 on LLaMA-3-8B. A pass is a block whose update
agrees with one earlier expert, but sharing it is only worth a separate expert when enough such blocks fall in the
same layer, so isolated agreements are left to the new task's own copy.

\paragraph{Direction test strictness:} At $\varphi$=0.006 (Table~\ref{tab:dir-ablation}), loosening the test from
strict ($Z$=3, $\tau$=+0.02) to normal ($Z$=2, $\tau$=0) raises the passing blocks from 1,446 to 3,226 and the shared
blocks from 468 to 1,074, but 31.2\% of the passing blocks barely agree in direction (cosine below 0.02). Relaxing
further adds little (3,851 passing, 1,182 shared), and 37.7\% of its passing blocks fall below that cosine. Even the
default test ($Z$=3, $\tau$=0) passes 33.0\% of its blocks with a cosine below 0.02, and 74\% of the blocks its sign
condition rejects have a cosine above $-0.02$, so near zero the sign condition separates blocks by noise rather than
by agreement. Clear agreement appears only at the second task, where two classification tasks meet: there 0.6 to
0.8\% of the passing blocks fall below a cosine of 0.02, against 30 to 53\% at later tasks. The size floor limits
sharing more than the direction test: it undoes about two thirds of the passing blocks at $\varphi$=0.006 and 92.4\%
at the default $\varphi$=0.012. OP is the same for all three tests (34.65 to 34.74), although the relaxed test
shares 2.5 times as many blocks, so on this stream the extra shared blocks neither help nor hurt.

\subsection{Minimum Expert Size}
\label{app:min_size}
The size floor of Eq.~\eqref{eq:sos_min_size} is $T=\varphi N_{\text{base}}$. A task selects about
$N_{\text{base}}/L$ blocks per layer in a model with $L$ adapted layers, so $T$ equals a fraction $\varphi L$ of a
layer's selection, independent of $\beta$. On LLaMA-2-7B ($L$=32), $\varphi$=0.012 gives $T$=11.8, 38\% of the
30.7 blocks a task selects per layer, while $\varphi$=0.025 gives $T$=51.2, 80\% of 64.0. A layer is carved only
when its shared part reaches $T$ and each remainder is either empty or also reaches $T$; at 80\% this requires the
two tasks to agree on nearly the whole layer. No shared block formed at any task of the $\beta$=25\% run
(Figure~\ref{fig:reuse_panels}c), so that run trains isolated experts only, while the $\beta$=12\% run formed between
13 and 176 shared blocks per split. The same floor also drops every layer in which a task selects fewer than $T$
blocks, which at $\varphi$=0.025 removes 5 to 9 of the 32 layers per task. A split that leaves both a shared piece and a non-empty remainder needs at
least $2T$ blocks in the layer, that is, $\varphi L \le 0.5$. With 36 layers, Qwen3-4B has $\varphi L$ of 0.43 and
0.90 for the two settings, so $\varphi$=0.012 satisfies this condition on both backbones and $\varphi$=0.025
violates it on both. At $\beta$=12\%, Qwen3-4B selects about 19 blocks per layer, so $T$=8.3 is 43\% of a layer's
selection and the floor already drops 2 to 6 of its 36 layers per task. The 25\% runs therefore differ from the
12\% runs in more than the budget: they have more blocks per adapted layer, but fewer adapted layers and no shared
experts, which is consistent with the different effect of $\beta$=25\% on forgetting across backbones.

\section{Block Selection}
\label{app:analysis}
This section analyzes SETA's block selection on TRACE-8 for Qwen3-4B, LLaMA-2-7B and LLaMA-3-8B at $\beta$=12\%:
which projection and layers receive the selected blocks, and how much each task overlaps with earlier ones.

\subsection{Cross-Task Block Overlap}
\label{app:phase0:overlap}

\begin{figure}[htbp]
  \centering
  \includegraphics[width=\linewidth]{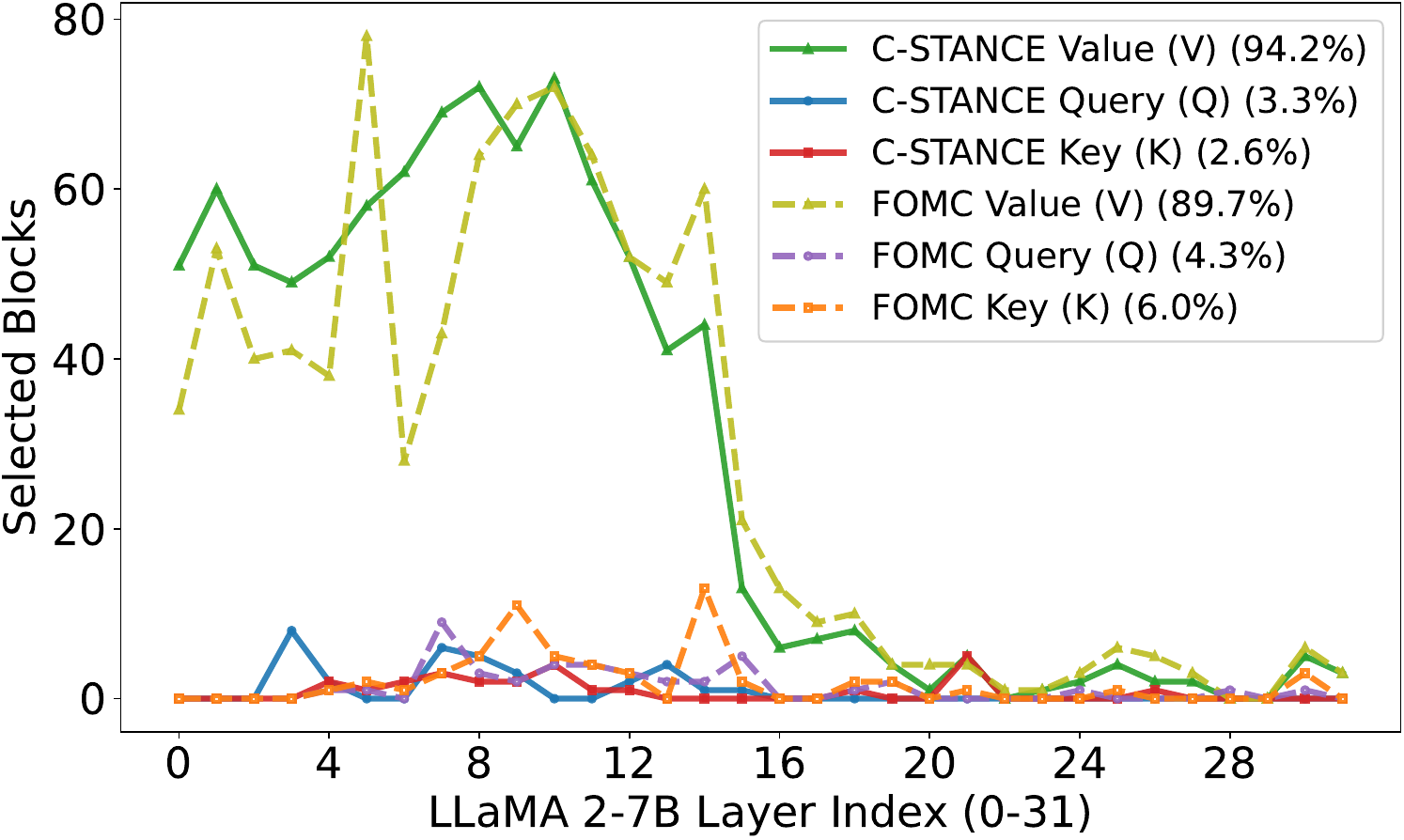}
  \caption{Layer-wise share of top-ranked gradients in the Q, K and V projections. Nearly all high-magnitude
  gradients fall in the Value projection.}
  \label{fig:gradient_analysis}
\end{figure}

On Qwen3-4B, the Jaccard overlap between the blocks of consecutive tasks ranges from 0.13 (NumGLUE-ds and
20Minuten) to 0.30 (C-STANCE and FOMC).

\subsection{Layer-wise Selection Across Architectures}
\label{app:phase0:layer}
\begin{figure*}[t]
  \centering
  \includegraphics[width=\textwidth]{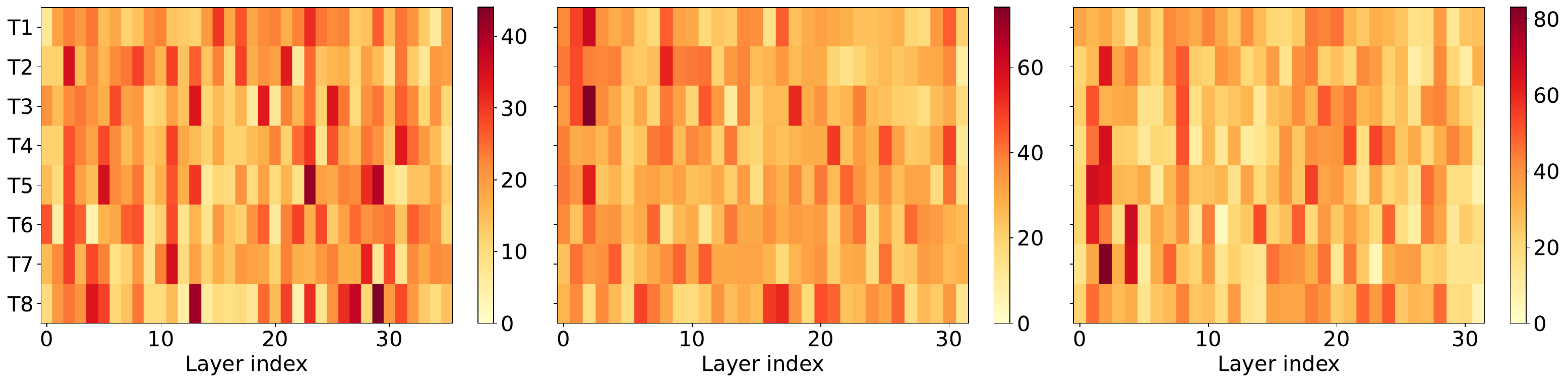}
  \caption{Selected blocks per layer and task on TRACE-8 for Qwen3-4B (left), LLaMA-2-7B (middle) and LLaMA-3-8B
    (right); darker cells mark more blocks. All three backbones spread their selection over every layer with a
    task-dependent profile.}
  \label{fig:layer_coverage}
\end{figure*}
Figure~\ref{fig:layer_coverage} shows that all three backbones spread their selection over all layers with a
non-uniform, task-dependent profile, without architecture-specific tuning.

\begin{figure*}[t]
  \centering
  \includegraphics[width=\textwidth]{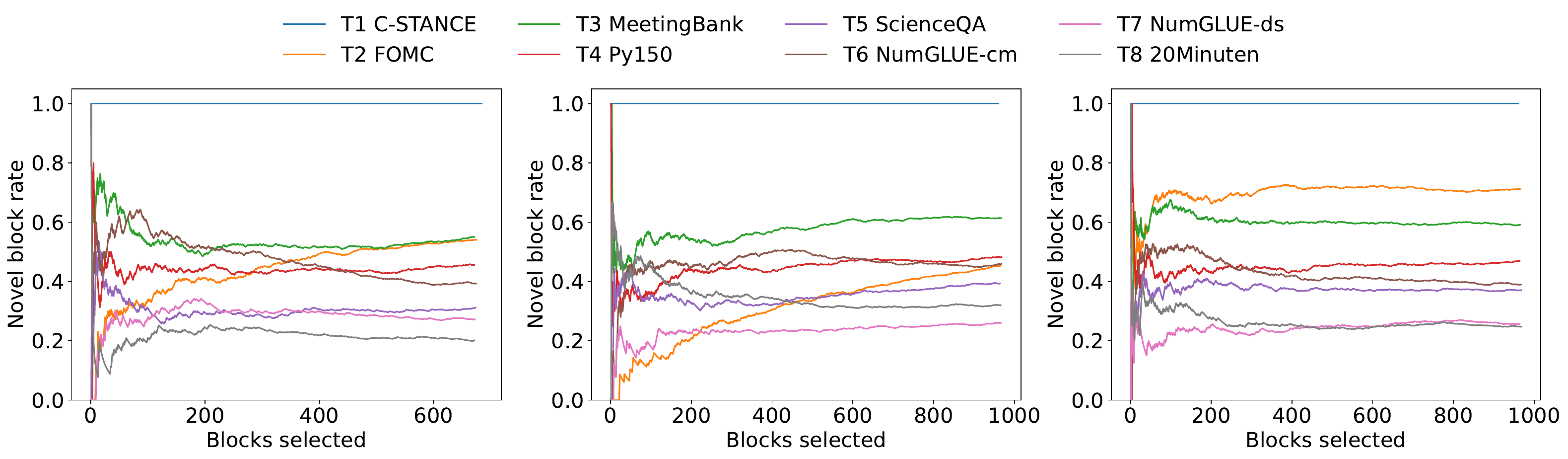}
  \caption{Rate of novel blocks during selection for Qwen3-4B (left), LLaMA-2-7B (middle) and LLaMA-3-8B (right):
    the fraction of the first $n$ selected blocks that no earlier task selected. The rate falls over the sequence
    as later tasks reuse blocks selected before them.}
  \label{fig:novel_rate}
\end{figure*}

\subsection{Sub-linear Growth of the Adapted Blocks}
\label{app:phase0:growth}
\begin{figure*}[t]
  \centering
  \includegraphics[width=\textwidth]{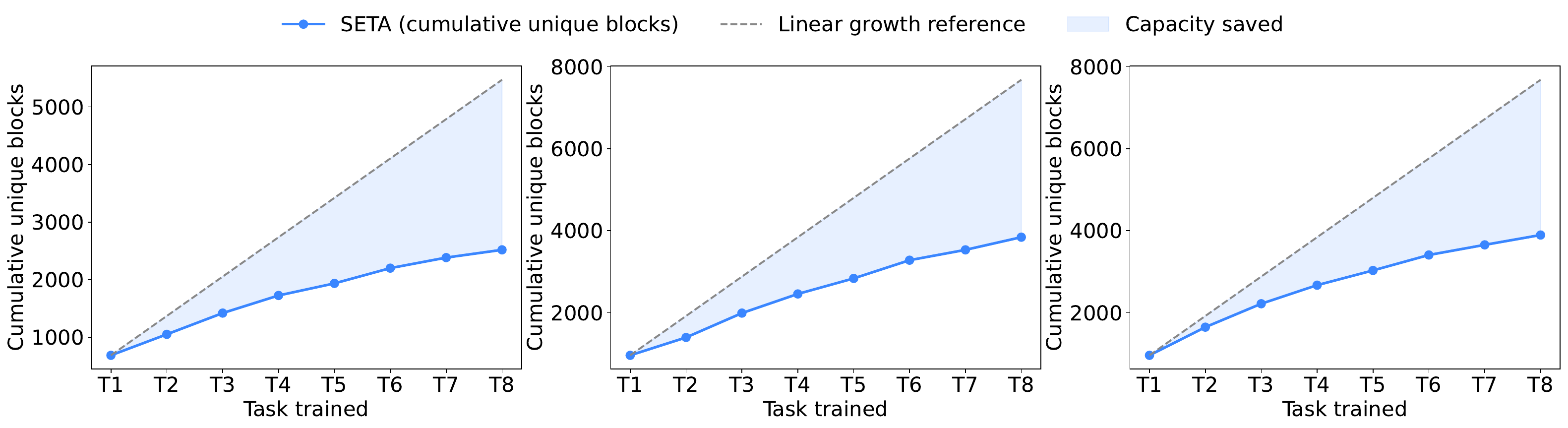}
  \caption{Distinct V-projection blocks adapted by any task so far on TRACE-8 ($\beta$=12\%) for Qwen3-4B (left),
    LLaMA-2-7B (middle) and LLaMA-3-8B (right), against the count if every task took blocks no earlier task had
    selected (dashed). Reuse keeps the adapted set growing more slowly than the number of tasks.}
  \label{fig:cumulative_models}
\end{figure*}
Each task reuses part of the blocks selected before it, so the set of adapted positions grows more slowly than the
number of tasks (Figure~\ref{fig:cumulative_models}). After eight tasks, Qwen3-4B with $\beta$=12\% covers 2,517
distinct blocks, 46.7\% of the 5,391 a non-reusing selection would need, and the blocks added per task fall from 684
for the first task to 135 for the last. This counts positions in the backbone, not stored parameters: a block
selected again by a later task is stored once more in the new task's expert unless it moves into a shared expert,
so the stored experts grow almost linearly.

\subsection{Novel Block Discovery per Layer}
\label{app:phase0:unique}
\begin{figure*}[t]
  \centering
  \includegraphics[width=\textwidth]{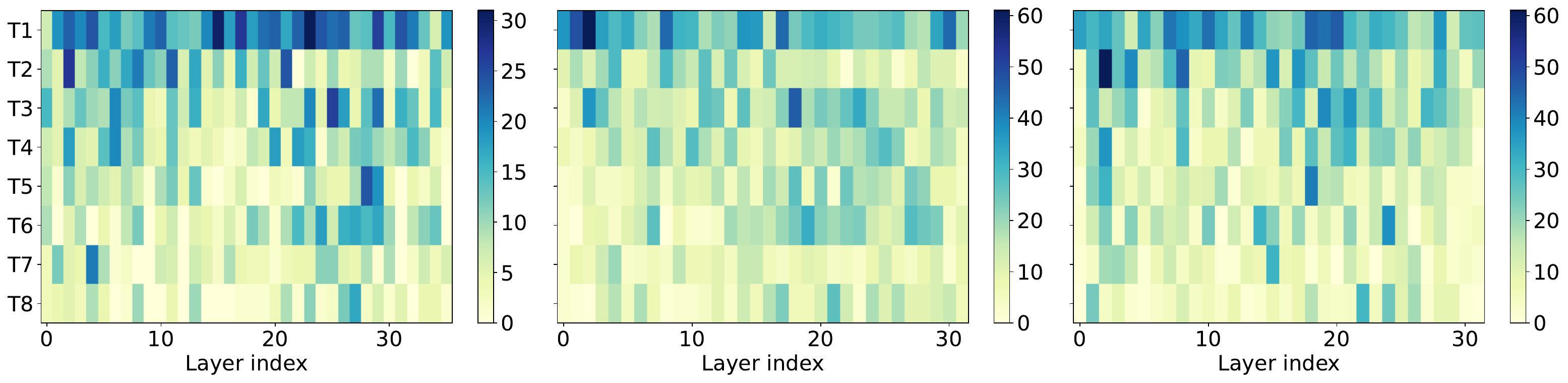}
  \caption{Blocks per layer and task that no earlier task selected at the same position, for Qwen3-4B (left),
    LLaMA-2-7B (middle) and LLaMA-3-8B (right); darker cells mark more new coverage. Later tasks add less new
    coverage as more layers are already covered.}
  \label{fig:unique_layer_heatmap}
\end{figure*}
Figure~\ref{fig:unique_layer_heatmap} resolves the novel-block rate of Figure~\ref{fig:novel_rate} by layer and is the
layer-level counterpart of the cumulative growth in Figure~\ref{fig:cumulative_models}. The first task fills
every layer, and later rows grow lighter on all three backbones as more layers are already covered.

\begin{table}[H]
\centering
\caption{Sensitivity to the split-rule settings on TRACE-8 with Qwen3-4B under the reduced protocol. Each row changes one setting from its default: the agreement bar $Z$, the size floor $\varphi$ or the kept
fraction share$_q$ of a shared piece. $\Delta$OP is the change from the full-system run of the same round (first
row), whose sign-test setting was not recorded. Only a size floor larger than a typical layer's selection breaks
the method. $^\dagger$Run with the sign test, so no $\Delta$OP is given. The share$_q$=0.5 run lacks one final
score, so only its LA is reported.}
\label{tab:abl_knobs_rule}
\small
\setlength{\tabcolsep}{3.5pt}
\begin{tabular}{lcccc}
\toprule
Setting & LA & OP & BWT & $\Delta$OP \\
\midrule
Full system & 41.73 & 34.29 & $-$8.50 &  \\
\midrule
$Z$=1 & 39.98 & 34.78 & $-$5.95 & +0.49 \\
$Z$=2 & 40.79 & 37.17 & $-$4.14 & +2.87 \\
$Z$=4 & 42.23 & 37.75 & $-$5.12 & +3.46 \\
\midrule
$\varphi$=0.006 & 41.76 & 35.91 & $-$6.69 & +1.61 \\
$\varphi$=0.025 & 39.44 & 33.64 & $-$6.62 & $-$0.65 \\
$\varphi$=0.04 & 35.57 & 28.56 & $-$8.02 & $-$5.74 \\
\midrule
share$_q$=0.5 & 40.86 &  &  &  \\
share$_q$=0.75$^\dagger$ & 40.89 & 37.13 & $-$4.30 &  \\
\bottomrule
\end{tabular}
\end{table}
\section{Ablations}
\label{app:ablation}
Tables~\ref{tab:abl_main} and~\ref{tab:abl_reduced} in Section~\ref{sec:exp_abl}
give the component ablations; this section discusses them and the sensitivity to one setting at a time.

\begin{figure*}[t]
  \centering
  \includegraphics[width=0.9\textwidth]{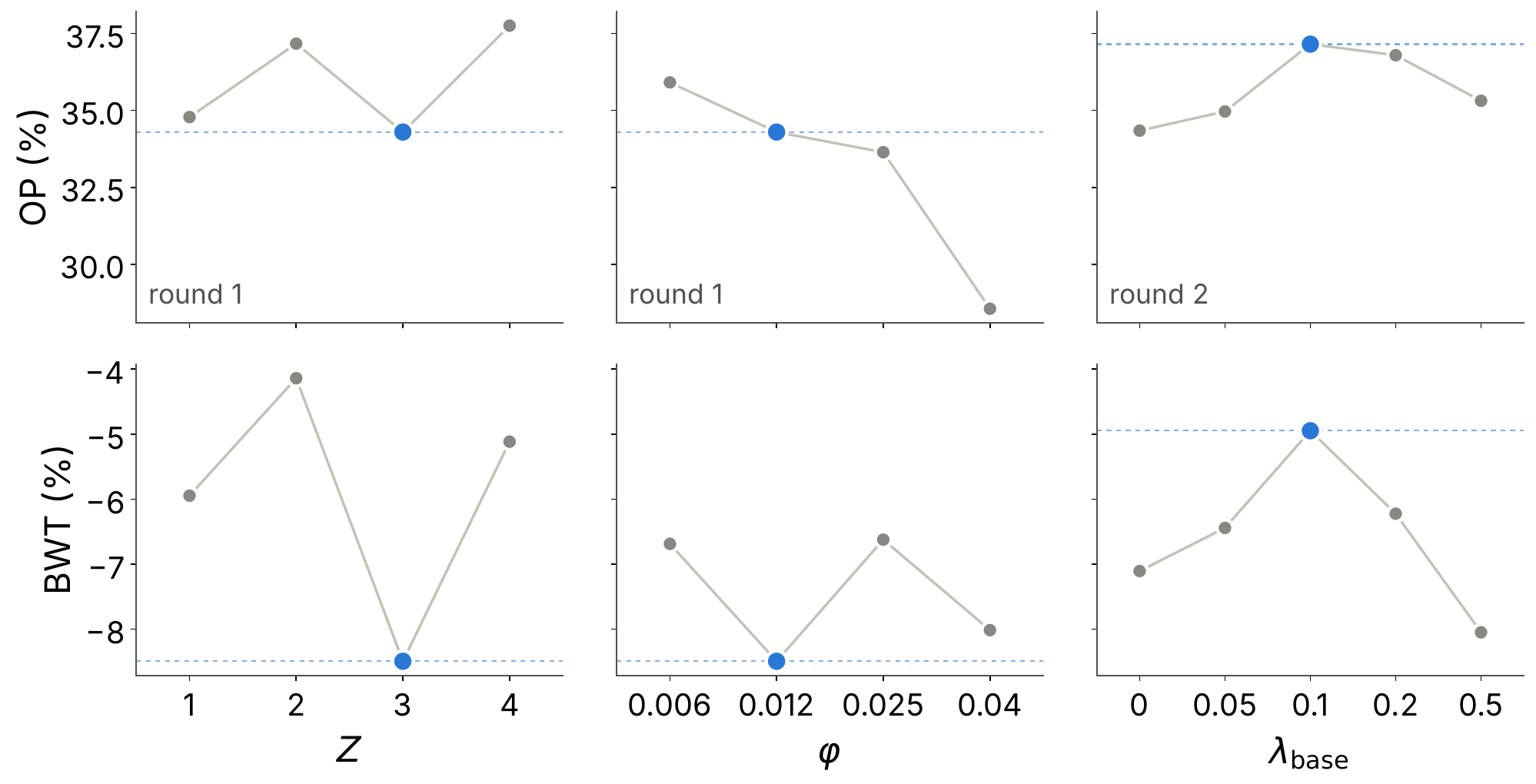}
  \caption{Sensitivity to one setting at a time on TRACE-8 with Qwen3-4B under the reduced protocol: OP
  (top) and BWT (bottom) against the agreement bar $Z$, the size floor $\varphi$ and the anchoring strength
  $\lambda_{\text{base}}$. Blue marks the default and the dashed line its value. The $Z$ and $\varphi$ runs share
  the full-system run of Table~\ref{tab:abl_knobs_rule}, and the $\lambda_{\text{base}}$ runs that of
  Table~\ref{tab:abl_reduced}; share$_q$ is not drawn. Only a size floor that is too large breaks the method.}
  \label{fig:sensitivity}
\end{figure*}

\paragraph{Sensitivity:}
Tables~\ref{tab:abl_knobs_rule}, and~\ref{tab:abl_knobs_lam} and Figure~\ref{fig:sensitivity}
use the reduced protocol of Table~\ref{tab:abl_reduced} (500 training samples), so their scores are compared only
within the protocol. The runs come in two rounds: round 1 holds the split-rule settings and their own full-system
run, and round 2 the component removals, the anchoring sweep and the direction-test arms, all with the sign test on.
The two full-system runs reach OP of 34.29 and 37.15, and whether the first used the sign test is not recorded, so
differences of a few points between single runs should not be read as rankings.

Only $\varphi$=0.04 breaks the method ($-5.74$ OP), because the floor then exceeds a typical layer's selection and
most layers are not adapted. The other split-rule settings move OP by $-0.65$ to $+3.46$ points, and the full-system
run scores lowest among the $Z$ settings, so they show only that no setting in this range breaks the method. For
anchoring, the default $\lambda_{\text{base}}$=0.1 gives the highest OP and the least forgetting of the five
strengths tested. Weaker anchoring forgets more, consistent with shared experts drifting toward later tasks, while
the strongest (0.5) learns most (LA 42.35) but also forgets most (BWT $-8.05$).

\section{Baseline Descriptions}
\label{app:baseline}
We compare SETA with LoRA and MoE based continual learning methods and one frozen reference. All LoRA baselines are trained with the code base of TreeLoRA~\cite{qian2025treelora} on the same task streams, data and evaluation as
SETA, with the settings of Table~\ref{tab:hyperparams}.

\begin{table*}[t]
\centering
\small
\caption{Training settings of SETA and the LoRA baselines (TreeLoRA code base). All methods train on the same task streams and data and use the same prompt format.}
\label{tab:hyperparams}
{\setlength{\tabcolsep}{6pt}\setlength{\aboverulesep}{0pt}\setlength{\belowrulesep}{0pt}%
\renewcommand{\arraystretch}{1.2}%
\begin{tabular}{p{3.1cm}|p{6.3cm}|p{4.4cm}}
\toprule
Setting & SETA & LoRA baselines \\
\midrule
Backbones & \multicolumn{2}{p{11.4cm}}{LLaMA-2-7B, LLaMA-3-8B, Qwen3-4B; the same local weights for all methods} \\
Prompt & \multicolumn{2}{p{11.4cm}}{Instruction prompt of Appendix~\ref{app:baseline}, without the model's chat template} \\
Training data & \multicolumn{2}{p{11.4cm}}{5,000 samples per task.} \\
Evaluation & \multicolumn{2}{p{11.4cm}}{First 500 test samples per task.} \\
Max.\ sequence length & \multicolumn{2}{p{11.4cm}}{512 tokens; MeetingBank 2,048 and 20Minuten 768 (1,536 and 512 on LLaMA-3-8B)} \\
\midrule
Learning rate & $9.65\times10^{-6}$ & $9.65\times10^{-6}$ \\
Schedule & Linear decay, 2 warm-up steps & Cosine decay, no warm-up \\
Optimization & AdamW (Trainer default), gradient clipping at 1.0 & DeepSpeed ZeRO-2 \\
Epochs per task & 3 & 3 \\
Effective batch size & 8 (2 $\times$ 4; 1 $\times$ 8 for MeetingBank, and for 20Minuten on LLaMA-3-8B) & 8 (1 $\times$ 8) \\
Weight decay & 0 & 0 \\
Trained parameters & V-projection blocks, $\beta$=12\% (25\% in the larger-budget runs) & LoRA adapters, Rank 8 \\
Regularization & Elastic anchoring, $\lambda_{\text{base}}$=0.1 & TreeLoRA: 0.5 \\
Seeds & 100, 42 & 100, 42 \\
Hardware & One NVIDIA A100 per run & One A40, A100 \\
\midrule
Block size & $256\times256$ (LLaMA-2-7B), $128\times128$ (LLaMA-3-8B, Qwen3-4B) & -- \\
Warm-up for block scoring & At most 100 steps, stopped dynamically (fixed 50 in the reduced protocol); learning rate $9.65\times10^{-6}$, $2\times10^{-6}$ for MeetingBank and 20Minuten & -- \\
Direction test & $Z$=3; sign test, $\tau$=0; 90\% energy, rank $\le$ 8 & -- \\
Size floor $\varphi$ & 0.012 (0.025 at $\beta$=25\%) & -- \\
Gating & Sigmoid per expert; new-expert bias 1.0; all gate rows trainable & -- \\
Expert dropout & 0 & -- \\
\bottomrule
\end{tabular}}
\end{table*}

\paragraph{FIX:} The pretrained backbone is not trained and answers each task from in-context examples. It
measures what the backbone achieves without continual training and is independent of the task order.

\paragraph{SeqLoRA:} A single LoRA adapter is fine-tuned on the tasks in sequence without any mechanism against
forgetting.

\paragraph{O-LoRA~\cite{wang2023orthogonal}:} Each task is learned in its own low-rank subspace, constrained to be
orthogonal to the subspaces of earlier tasks.

\paragraph{HiDe-LoRA~\cite{wang2024hide}:} The continual learning objective is decomposed hierarchically into
within-task prediction, task-identity inference and task-adaptive prediction, with task-specific and task-shared
LoRA components.

\paragraph{TreeLoRA~\cite{qian2025treelora}:} Layer-wise LoRA adapters are organized in a tree built from the
gradient similarity between tasks, which decides how a new task reuses the adapters of earlier ones.

\paragraph{MoE-CL~\cite{kang2025moecl}:} Each task trains its own LoRA expert, together with a shared LoRA
expert that is trained against a task discriminator so that it keeps features common to the tasks. It is run
with its released code, using the learning rate and number of epochs of SETA.
\subsection{Prompt Design for Training and Inference}

To clarify, all models are utilized in a strictly generative text-to-text paradigm via causal language modeling. We do not append any task-specific classification heads to the architecture. Both training and inference are conducted using a standardized instruction-tuning prompt format. Every data instance across all tasks is uniformly structured with a system prefix, the task instruction including context, and a response trigger:

\begin{verbatim}
Below is an instruction that describes 
a task. 
Write a response that appropriately 
completes the request.

### Instruction:
{task_specific_instruction_and_input}

### Response:
{target_label_or_text}
\end{verbatim}

During evaluation, the model is provided the prompt up to the response trigger, and the output is generated autoregressively. Because we do not use a classification head, we map the raw generated text back to the discrete labels using dataset-specific parsing scripts. For classification tasks, the parsing functions extract the predicted option or numeric label from the generated string and compare it against the target ground truth to compute accuracy. Py150 is scored by the edit similarity between the generated and the reference code. For generation tasks, the raw generated text is directly evaluated against the reference: MeetingBank uses ROUGE-L F1 against the reference summary, while 20Minuten uses SARI computed from the source sentence, the generated text, and the reference, normalized to the 0 to 100 range. These scores are averaged with the classification accuracies when computing OP.
\section{Detailed Dataset Specifications}
\label{app:datasets}
\paragraph{TRACE-8.} The eight tasks come from TRACE~\cite{wang2023trace} and are trained in this order:
C-STANCE, Chinese stance detection~\cite{zhao2023c}; FOMC, hawkish/dovish classification of monetary-policy
statements~\cite{shah2023trillion}; MeetingBank, meeting summarization~\cite{hu2023meetingbank}; Py150, Python code
completion~\cite{lu2021codexglue}; ScienceQA, science question answering~\cite{lu2022learn}; NumGLUE-cm and
NumGLUE-ds, arithmetic word problems~\cite{mishra2022numglue}; and 20Minuten, German text
simplification~\cite{kew202320}. MeetingBank is scored by ROUGE-L, Py150 by edit similarity, 20Minuten by SARI, and
the other tasks by accuracy. Training uses 5,000 samples per task, and
evaluation 500 test samples.
\paragraph{O-LoRA benchmark.} Orders 1--3 follow \citet{wang2023orthogonal} on four text-classification
datasets~\cite{zhang2015character}: order 1 is DBpedia, Amazon, Yahoo, AG News; order 2 is DBpedia, Amazon, AG News,
Yahoo; order 3 is Yahoo, Amazon, AG News, DBpedia. Orders 4--6 follow \citet{razdaibiedina2023progressive} on fifteen
tasks from GLUE~\cite{wang2018glue}, SuperGLUE~\cite{wang2019superglue}, IMDB~\cite{maas2011learning} and the
classification datasets above. Order 4 is MNLI, CB, WiC, COPA, QQP, BoolQ, RTE, IMDB, Yelp, Amazon, SST-2,
DBpedia, AG News, MultiRC, Yahoo; order 5 is MultiRC, BoolQ, WiC, MNLI, CB, COPA, QQP, RTE, IMDB, SST-2, DBpedia,
AG News, Yelp, Amazon, Yahoo; order 6 is Yelp, Amazon, MNLI, CB, COPA, QQP, RTE, IMDB, SST-2, DBpedia, AG News,
Yahoo, MultiRC, BoolQ, WiC. Each task uses 500 training and 100 test samples.

\paragraph{Warm-up.} The dynamic warm-up of Section~\ref{sec:sparse_construction} runs between 15 and 100 steps,
estimates the statistics of $\rho(t)$ over a trailing window of $W$=6 steps, and stops after $S$=10 consecutive
steps below the threshold, or $S$=15 for MeetingBank and 20Minuten. 

\section{Computational Overhead}
Let $d$ be the hidden dimension, $l$ the block size, $E_l$ the number of experts that hold blocks in an adapted
layer and $n_l$ the number of selected blocks in that layer. In addition to the frozen product $W_0\mathbf{x}$,
SETA computes one gate logit per expert, at a cost of $\mathcal{O}(dE_l)$ per token, and a block-sparse product
over the selected blocks, at a cost of $\mathcal{O}(n_l l^2)$ per token. All experts are evaluated at inference,
since every expert has its own sigmoid gate (Section~\ref{sec:sigmoid_gating}); Table~\ref{tab:routing_avgs} shows
that keeping only the top-$k$ gates lowers the average score.
\section{SETA Algorithm}
Algorithm~\ref{alg:seta_algo} details the SETA training procedure.
For each task $t$, Line~3 selects the sparse blocks $\mathcal{P}_t$ with the warm-up and scoring pipeline of
Section~\ref{sec:sparse_construction}. For $t>1$, Lines~7--10 apply Split-on-Share layer by layer: overlapping
blocks that pass the direction test and the size rule form the shared expert $E_{\text{s}}$, the remainder of the
prior expert stays frozen in $E_{\text{u}}^{(t-1)}$, and the new task's own blocks initialize $E_{\text{u}}^{(t)}$.
Prior unique experts are frozen in Line~13, the gating network is expanded in Line~14, and the model is optimized
in Line~15 over $\mathcal{L}_{\text{task}} + \mathcal{L}_{\text{reg}}$. Line~16 snapshots the shared expert weights
before the next task begins.

\label{app:algorithm}
\begin{algorithm}[h]
\caption{SETA: Sparse Subspace-to-Expert Training Procedure}
\label{alg:seta_algo}
\begin{algorithmic}[1]
\REQUIRE Pre-trained $W_0$, Task Sequence $\mathcal{T} = \{T_1, \dots, T_k\}$,
         block budget $\beta$, size floor $\varphi$,
         Regularization coefficient $\lambda_{\text{base}}$
\STATE \textbf{Initialize:} $\mathcal{E} \leftarrow \emptyset$,
       $W_g \leftarrow \emptyset$
\FOR{$t = 1$ to $k$}
    \STATE \textbf{Subspace Selection:} Identify active sparse blocks
           $\mathcal{P}_t$ via adaptive gradient pipeline.
    \IF{$t=1$}
        \STATE Construct initial unique expert $E_{\text{u}}^{(1)}$
               on $\mathcal{P}_1$.
    \ELSE
        \FOR{each layer $l \in L$}
            \STATE \textbf{SoS Decomposition:} Compute $\mathcal{I}_l$, $\mathcal{R}_l$ and $\mathcal{N}_l$
                   with the direction test (Section~\ref{sec:direction_sharing}) and the size rule
                   (Eq.~\ref{eq:sos_min_size}).
            \STATE \textbf{Expert Allocation:} $E_{\text{s}} \leftarrow
                   \mathcal{I}_l$ via weight inheritance;
                   freeze $E_{\text{u}}^{(t-1)} \leftarrow \mathcal{R}_l$.
            \STATE \textbf{New Expert:} Initialize $E_{\text{u}}^{(t)}$
                   on $\mathcal{N}_l$.
        \ENDFOR
    \ENDIF
    \STATE \textbf{Freeze History:} $\nabla E_{\text{u}}^{(k)} = 0$
           for all $k < t$.
    \STATE \textbf{Gating Expansion:} Resize $W_g$ to $|\mathcal{E}|$;
           preserve logit invariance.
        \STATE \textbf{Optimize:} Minimize $\mathcal{L}_{\text{task}} + \mathcal{L}_{\text{reg}}$
       over $E_{\text{u}}^{(t)}$, $E_{\text{s}}$ and all gate rows of $W_g$
       (Eq.~\ref{equ:seta_train_object}).
    \STATE \textbf{Anchor Update:} $\hat{W}_i \leftarrow W_i$ for all
           shared experts.
\ENDFOR
\STATE \textbf{Inference:} Task-agnostic retrieval via $W_g$
       (Eq.~\ref{eq:task_agnostic_pred}).
\end{algorithmic}
\end{algorithm}

\end{document}